\documentclass[letterpaper]{article} 
\usepackage{aaai2027}  
\usepackage[hyphens]{url}  
\usepackage{graphicx} 
\usepackage{natbib}  
\usepackage{caption} 
\usepackage{algorithm}
\usepackage{algorithmic}

\usepackage{newfloat}
\usepackage{listings}
\DeclareCaptionStyle{ruled}{labelfont=normalfont,labelsep=colon,strut=off} 
\floatstyle{ruled}
\newfloat{listing}{tb}{lst}{}
\floatname{listing}{Listing}

\usepackage{booktabs}

\usepackage{multirow}
\usepackage{makecell}
\usepackage{xcolor}
\usepackage{colortbl}
\usepackage{tikz}
\usepackage{amsmath}
\usepackage{amssymb}

\definecolor{falseNeg}{HTML}{3d9af5}
\definecolor{falsePos}{HTML}{fa5757}

\definecolor{gold}{HTML}{FBF2D2}
\definecolor{silver}{HTML}{DDDDDD}
\definecolor{bronze}{HTML}{EED2B8}

\definecolor{goldD}{HTML}{D9AE13}
\definecolor{silverD}{HTML}{909090}
\definecolor{bronzeD}{HTML}{9A5F26}

\newcommand{\medal}[3]{\tikz[baseline=(char.base)]{\node[rounded corners=2pt,fill=#1,draw=#2,inner sep=1pt](char){#3};}}

\newcommand{\bm}[2]{
    \ifcase#1\or
      {\medal{gold}{goldD}{\textbf{#2}}}
    \or 
      {\medal{silver}{silverD}{#2}}
    \or 
      {#2}
    \else 
      #2
    \fi\ignorespaces
}

\definecolor{weights}{HTML}{0398fc}

\let\titleold\title
\renewcommand{\title}[1]{\titleold{#1}\newcommand{\thetitle}{#1}}
\def\maketitlesupplementary
   {
   \newpage
        \twocolumn[
        \centering
        \Large
        \textbf{\thetitle}\\
        \vspace{0.5em}Supplementary Material \\
        \vspace{1.0em}
       ]
   }

\usepackage{cleveref}

\title{ChangeFlow - Generative Remote Sensing Change Detection Using Latent Rectified Flow}
\author{
    Blaž Rolih,
    Matic Fučka,
    Filip Wolf,
    Luka Čehovin Zajc
}
\affiliations{
   University of Ljubljana, Faculty of Computer and Information Science, Slovenia\\
     blaz.rolih@fri.uni-lj.si
}

\nocopyright

\begin{document}

\maketitle

\begin{abstract}
Remote sensing change detection (RSCD) localises changes between two images of the same geographic region. Most state-of-the-art methods are trained with a per-pixel discriminative objective that classifies each spatial location independently.  In this scenario, the predicted changed region is not modelled as a coherent whole, so predictions tend to be spatially fragmented. Generative modelling offers a principled solution: by learning a distribution over plausible change masks, it treats the mask as a single object and encourages global consistency. Yet existing generative RSCD methods lag behind strong discriminative baselines, held back by costly pixel-space generation and overly complex conditioning. We introduce \textbf{ChangeFlow}, which reformulates change detection as the generative synthesis of change masks in a compact latent space via rectified flow, guided by a structured yet lightweight bi-temporal conditioning signal. Changeflow yields spatially coherent predictions without sacrificing efficiency: across four binary benchmarks, SYSU, LEVIR, CLCD, and OSCD, ChangeFlow achieves an average F1 of $80.4\%$, a $1.3$-point gain over the previous best with better efficiency. It also extends to semantic change detection, setting a new state-of-the-art $65.9$ $F_{scd}$ on SECOND. Project page: \url{https://blaz-r.github.io/changeflow_cd}
\end{abstract}


\section{Introduction}
\label{sec:intro}

Remote sensing change detection (RSCD) aims to localise changes between two (or more) images of the same geographic region acquired at different times~\cite{daudt2018fcn, chen2021bit}. With the increasing availability of high-resolution remote sensing imagery and advances in deep learning, RSCD has become a key component in applications such as environmental monitoring, land-use mapping, disaster response, and urban development~\cite{hansch2024eo4climate, zhu2022rsLandChange}. 

Annotated changes are represented as pixel-wise maps, but the underlying target is not just a collection of independent pixels. 
Changed regions delineate coherent changed objects and boundaries; consequently, change masks exhibit strong spatial structure. 
Most recent RSCD methods, however, follow a dense discriminative paradigm: they predict changes at each spatial location and are supervised with objectives on individual pixels~\cite{rolih2025btc, cheng2025changedino}.
Although modern architectures aggregate rich spatial context through convolutional and attention-based methods, they do not directly model the mask itself as a joint structured object~\cite{yu2024maskcd}. This mismatch can lead to fragmented predictions and poor global consistency.

\begin{figure}[!t]
    \centering
    \includegraphics[width=1\linewidth]{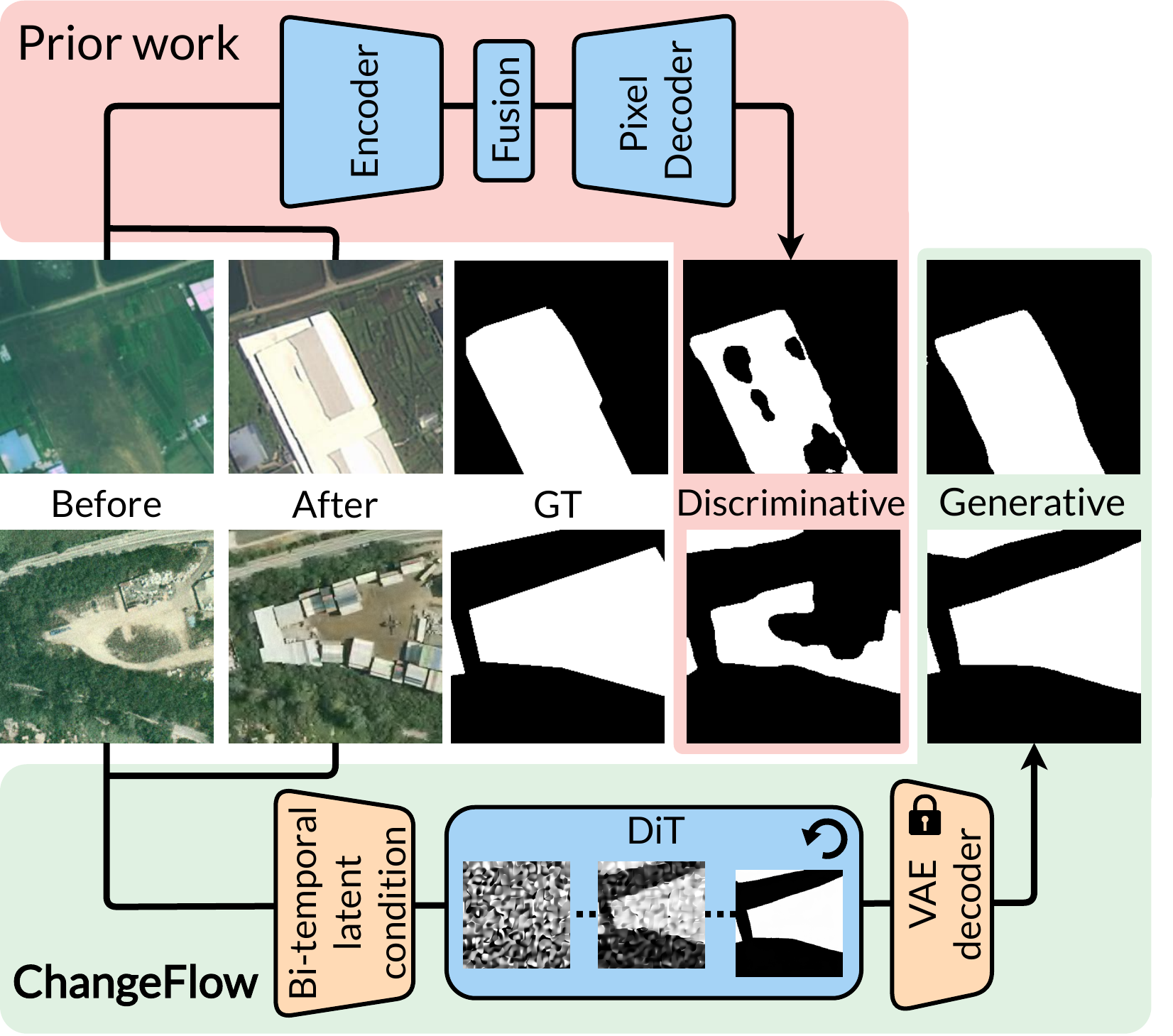}
    \caption{Change masks are represented as pixel grids, but annotators delineate coherent changed objects and regions. Dense discriminative methods supervise the mask through pixel-wise labels and capture global structure only indirectly. \textbf{ChangeFlow}, instead, generates the mask in the latent space, modelling it as a single structured entity and encouraging globally coherent predictions.}
   \label{fig:hostpot}
\end{figure} 

Instead of treating the mask as a set of individual pixels, a natural way to address this limitation is to model the change mask jointly, i.e., by sampling it from a conditional mask distribution.
Generative models provide such a framework: rather than learning a decision boundary between changed and unchanged pixels, they model the distribution of plausible outputs conditioned on the pair of input images. In principle, this allows for the direct capture of long-range spatial dependencies and region-level consistency. Recent generative change detection methods have explored this direction~\cite{jia2024smdnet, wen2024gcd-ddpm}, but they still lag behind discriminative approaches. We argue that this gap largely stems from design choices inherited from RGB image generation that are ill-suited to change masks. In particular, operating in pixel space makes iterative generation unnecessarily expensive, while complex conditioning pipelines, such as auxiliary predictions or heavy attention modules, increase the difficulty of optimisation and reduce practicality.

To make generative change detection both effective and efficient, we introduce \textbf{ChangeFlow}, a latent-space rectified-flow framework for RSCD. ChangeFlow reformulates change detection as \textit{change mask synthesis in latent space}: a pretrained VAE first encodes ground-truth change masks into a compact latent representation, and a diffusion transformer (DiT) is trained with a rectified-flow objective~\cite{liu2023rectifiedflow} to transport Gaussian noise to mask latents. The generative process is conditioned on simple bi-temporal feature differences extracted from the input image pair, avoiding auxiliary predictors and elaborate conditioning modules. This design keeps the model practical while preserving the key advantage of generative modelling.

Crucially, such a formulation yields several useful properties inherent to the model itself that discriminative models would require additional components to achieve (e.g., loss design).
First, because ChangeFlow jointly generates the mask, it encourages spatially coherent predictions with fewer holes, which we validate quantitatively. Second, the number of rectified-flow sampling steps is adjustable at inference time, providing a speed-accuracy trade-off without retraining. Third, because inference starts from noise, the model can sample multiple plausible masks for the same image pair, enabling the aggregation of multiple generated masks into a single robust prediction. This sampling-based view remains underexplored in segmentation rectified-flow models~\cite{wang2024semflow}.

In summary, our contributions are as follows: 
\begin{itemize}
    \item As our main contribution, we reformulate RSCD as latent-space change mask generation and propose a rectified flow framework that jointly models change masks and produces globally coherent predictions.
    \item We show that auxiliary predictors and complex layers are unnecessary for conditioning and propose a simpler strategy based on feature difference.
    \item We demonstrate the generality of the formulation by extending ChangeFlow to semantic change detection.
\end{itemize}

We validate ChangeFlow on four standard binary change detection (BCD) datasets: SYSU, LEVIR, CLCD, and OSCD. ChangeFlow achieves F1 scores of $85.6$\%, $92.1$\%, $84.5$\%, and $59.5$\%, respectively, outperforming previous methods on three datasets. This sets a new best average F1 of $80.4$\% across all four datasets, outperforming the previous-best, ChangeDino~\cite{cheng2025changedino}, by $1.3$ percentage points. We further extend ChangeFlow to semantic change detection (SCD), where it achieves a new state of the art on SECOND with 65.9 $F_{scd}$, improving over TaCo~\cite{guo2025taco} by $1$ point.

\section{Related work}

\noindent\textbf{Remote sensing change detection (RSCD).} RSCD has evolved in recent years from pixel-wise differencing and statistical tests to end-to-end deep models~\cite{singh1989reviewCD, peng2025deepDCSurvey}. Since early deep models, the field relied on Siamese networks, from convolutional architectures~\cite{daudt2018fcn, chen2020levirStanet}, to more recent transformer variants~\cite{bandara2022changeFormer,zhang2022swinsunet, rolih2026mason}, state-space models~\cite{chen2024changeMamba} and diffusion-inspired designs for the backbone~\cite{bandara2025ddpmcd, wen2024gcd-ddpm}. Beyond architectural advances, large-scale pretraining is increasingly important for performance and robustness~\cite{rolih2025btc, li2024ban, cheng2025changedino, wolf2026deo}. Recent work also explores \emph{semantic} change detection, which predicts change together with semantic categories~\cite{benidir2025hyscdg, guo2025taco, ding2024scannet, chen2024changeMamba}. In all settings, the dominant formulation remains discriminative (pixel-wise changed/unchanged classification), which often trades robust change-region modelling for straightforward supervised training. We instead cast CD as an iterative generative inference problem that explicitly models the distribution of possible change masks and predicts the mask as a whole, thereby improving mask coherence.

\noindent\textbf{Generative models for computer vision tasks.} Generative models, particularly diffusion~\cite{nichol2021ddpm} and flow-based~\cite{liu2023rectifiedflow} formulations, have recently gained traction as powerful tools for visual representation learning. Such models were successfully applied to various fields, such as few-shot counting~\cite{vsuvstar2025codi}, anomaly detection~\cite{fuvcka2024transfusion}, monocular depth estimation~\cite{ke2024repurposing}, and object detection~\cite{chen2023diffusiondet}. Most relevant to our case, it has also been successfully applied to Earth Observation (EO) tasks (e.g., FlowEO~\cite{bellier2025floweo}) and to general semantic segmentation (e.g., SemFlow~\cite{wang2024semflow} and GSS~\cite{chen2023genSeg}). However, unlike ChangeFlow, such approaches rarely leverage the multiple-samples-based inference that generative models offer.

\noindent\textbf{Generative models for change detection.} Generative models have been used for CD in three ways. Most indirectly, they \textit{synthesise pseudo-changes} to enlarge and diversify the training set~\cite{zgeng2025changen2, song2024syntheworld, wang2024diffPseudo, benidir2025hyscdg, korkmaz2025referringCD}, acting as offline data generators. Others train diffusion models on remote sensing imagery and \textit{repurpose them as feature extractors}~\cite{bandara2025ddpmcd, jiang2025d3pm, jia2025satdifuser}, feeding the features to a discriminative head. In both, the change mask is still produced by a separately trained discriminative network. Closest to our work, a few methods formulate \textit{CD itself as generation}: GCD-DDPM~\cite{wen2024gcd-ddpm} conditions diffusion on the output of an auxiliary, attention-enhanced detector, and SMDNet~\cite{jia2024smdnet} integrates bi-temporal encodings into a pixel-space DDIM process. These operate in pixel space, require many generation steps, and rely on complex conditioning, which increases cost and limits performance. In contrast, ChangeFlow generates the mask directly in a compact latent space via rectified flow with lightweight feature-difference conditioning, enabling more efficient and accurate change-mask generation.

\section{Preliminaries}
\label{sec:rf_prelim}

Rectified flow (RF)~\cite{liu2023rectifiedflow} is a generative framework that maps Gaussian noise $X_0 \sim \mathcal{N}(0, I)$ to a target data distribution $X_1 \sim P_{data}$ via a straight-line trajectory. The intermediate state at any time $t \in [0, 1]$ is defined by linear interpolation:
\begin{equation}
X_t = (1-t)X_0 + tX_1 \enspace.
\end{equation}
Because this trajectory has a constant velocity of $(X_1 - X_0)$, a neural network $v_\theta(X_t, t)$ can be trained to predict it by minimising the mean squared error:
\begin{equation}
\min_{\theta} \mathbb{E}_{t, X_0, X_1} \left[ \| (X_1 - X_0) - v_\theta(X_t, t) \|^2 \right] \enspace ,
\end{equation}
where $t$ is sampled from $[0, 1]$. During inference, data is generated by integrating the predicted velocity field $v_\theta$ starting from a noise sample $X_0$. Unlike DDPMs, the straight path enables accurate integration in very few steps; RF is thus the standard in modern latent generation~\cite{esser2024sd}.

\begin{figure*}[t]
    \centering
    \includegraphics[width=\linewidth]{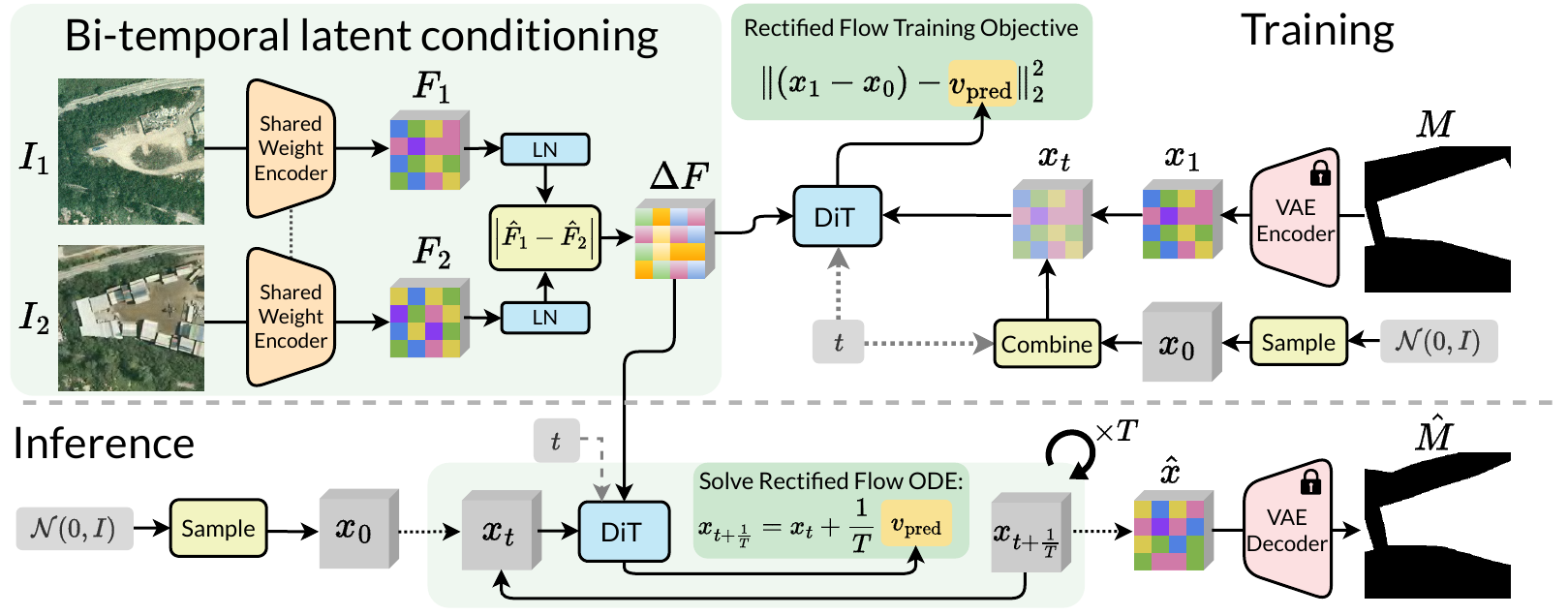}
    \caption{\textbf{Up.} Training pipeline of ChangeFlow using latent rectified flow conditioned on bi-temporal feature difference. \textbf{Down.} During inference, ChangeFlow iteratively generates a change mask by integrating the velocity field.}
    \label{fig:arch}
\end{figure*}

\section{ChangeFlow}

Recent attempts that use generative modelling for change detection disregard latent formulations, thereby increasing computational complexity. In contrast, we move our modelling process from the pixel to the latent space, adopt rectified flow for its few-step sampling, and use a principled conditioning scheme based on features extracted from a strong pretrained encoder.

Given a pair of images, we first extract features using a \textit{Shared Weight Encoder}, and we \textit{condition} the \textit{Diffusion Transformer (DiT)} rectified flow model on the absolute difference of the extracted features. Guided by this conditioning, the model then iteratively generates a latent representation of the corresponding change mask, which is ultimately decoded by the \textit{Variational Autoencoder (VAE)} into a change mask. The method is illustrated in Fig.~\ref{fig:arch} and described in detail in the following sections.

\subsection{Change Detection as Latent Generative Synthesis}
\label{sec:gen_cd}

\noindent\textbf{Change masks in latent space.} To explicitly model the distribution of change masks in latent space and obtain coherent predictions, we formulate change detection as a mask-generation problem. More specifically, we use rectified flow to generate change masks inside the latent space of a pretrained VAE~\cite{kingma2014vae}. While it is known that VAEs efficiently encode RGB images~\cite{esser2024sd, podell2024sdxl}, it is unclear whether this holds for binary images (i.e., change masks) and semantic change maps (in the case of SCD). To verify this, we perform a simple experiment and report the F1 score, border F1 score (BF1), and mean absolute error (MAE) in Tab.~\ref{tab:vae_pass}. We first repeat the binary change mask 3 times along the channel dimension, encode it with the SD-XL~\cite{podell2024sdxl} VAE, decode the resulting latent, and average the 3 output channels to restore the binary mask. The high F1 and BF1 scores, paired with a low MAE, indicate that this is indeed feasible and offers potential insights for applications beyond change detection.
\begin{table}[!t]
    \centering
    \caption{F1, border F1 (BF1), and mean absolute error (MAE) of binary ground truth masks reconstruction through SD-XL VAE~\cite{podell2024sdxl}. Pretrained VAE produces minimal information loss.}
    \centering
        \begin{tabular}{lcccc}
        \toprule
        & SYSU & LEVIR & CLCD & OSCD \\
        \midrule
             F1 $\uparrow$& 99.9& 99.3& 99.5& 99.4\\
             BF1 $\uparrow$ & 100.00& 100.00& 99.96& 99.99 \\
            MAE $\downarrow$ & 0.0004& 0.0007& 0.0006&0.0006\\
         \bottomrule
        \end{tabular}
    \label{tab:vae_pass}
\end{table}
We also verify that the encoding of semantic masks exhibits minimal information loss. We encoded and decoded the semantic masks using a VAE, as explained later in the paper, and calculated the SCD metrics $F_{scd}$ and $mIoU$, which remained at 99.45 and 99.53, respectively.

\noindent\textbf{Change mask rectified flow.} Let $M \in \{0,1\}^{H \times W}$ denote the binary ground-truth change mask (SCD encoding process is explained later in the paper) where $H$ and $W$ are the mask dimensions and $\mathcal{V}$ is a pretrained VAE encoder $\mathcal{V}$: $\mathbb{R}^3 \rightarrow \mathbb{R}^d$ (in our case SD-XL~\cite{podell2024sdxl} VAE). As described in the previous section, we can then encode the change mask with $\mathcal{V}$ by repeating the mask along the channel dimension:
\begin{equation}
    x_1 = \mathcal{V}(\{M, M, M\}), \quad x_1 \in \mathbb{R}^{h \times w \times d} \enspace.
\end{equation}
This yields a compact latent representation $h < H, w < W$.

During training, we sample Gaussian noise in the same shape as the latent space to obtain an \textit{initial state $x_0$}:
\begin{equation}
    x_0 \sim \mathcal{N}(0, I), \quad x_0 \in \mathbb{R}^{h \times w \times d} \enspace,
\end{equation}
which we use to construct an interpolated latent (i.e., an intermediate step along the straight trajectory) representation at a specified time step $t$:
\begin{equation}
    x_t = (1-t) x_0 + t x_1 \enspace.
    \label{eq:rf_interpolation}
\end{equation}
Following previous work~\cite{esser2024sd}, we sample timesteps in a logit-normal fashion, which emphasises learning at the critical point where $t=0.5$:
\begin{equation}
    t \in [0, 1]; \quad t = sigmoid(s); \quad s \sim \mathcal{N}(0, 1) \enspace.
\end{equation}
This represents the most ambiguous point in time at which the levels of noise and signal are balanced, with trajectories overlapping most, and the model must learn to rectify the field (see~\cite{esser2024sd} for more details).

To guide the network from initial noise to the final mask latent space, we prepare a bi-temporal latent conditioning signal $\Delta F$, which we will explain at the end of this subsection. We concatenate it with $x_t$ in the channel dimension and feed the resulting vector to the model. The rectified flow vector field is then parametrised using a DiT~\cite{peebles2022dit}-based network $\mathcal{M}_\theta$:
\begin{equation}
    v_{\text{pred}} = \mathcal{M}_\theta([x_t, \Delta F], t) \enspace.
\end{equation}
We train the network using the standard MSE loss for rectified flow~\cite{liu2023rectifiedflow}:
\begin{equation}
    \mathcal{L}_{\text{RF}} = \left\| (x_1 - x_0) - v_{\text{pred}} \right\|_2^2\enspace.
    \label{eq:rf_loss}
\end{equation}
This means that there is no explicit per-pixel objective; the model learns the velocity field at a specific time step (i.e., at a specific location along the straight trajectory). 
Because the objective models the mask latent jointly rather than as a product of discriminative per-pixel terms, the prediction is treated as a single coherent object by construction.
The process is also depicted at the top of Fig.~\ref{fig:arch}.

\noindent\textbf{Change mask generation guidance.}
To create a conditioning signal used to guide the generation process, we first extract high-level latent features from an image pair $(I_1, I_2)$ using a pretrained encoder $\Phi$ with shared weights:
\begin{equation}
    F_1 = \Phi(I_1), \quad
    F_2 = \Phi(I_2), \quad
    F_1, F_2 \in \mathbb{R}^{h' \times w' \times c}.
\end{equation}
To remain agnostic to temporal ordering and feature magnitude, we construct the conditioning signal as the absolute difference of the layer normalised (LayerNorm~\cite{ba2016layer} - LN) feature maps:
\begin{equation}
    \Delta F = \left| \mathrm{LN}(F_1) - \mathrm{LN}(F_2) \right| \enspace.
    \label{eq:dlatent}
\end{equation}
The process is also illustrated in the top-left of Fig.~\ref{fig:arch}. 
This approach offers an efficient latent design that enables strong conditioning for the task. Unlike previous generative change detection works~\cite{wen2024gcd-ddpm, jia2024smdnet}, it avoids complex auxiliary methods and attention-based conditioning, which are more prone to overfitting.

\subsection{Inference via Rectified Flow Integration}

At inference time, given a pair of images ($I_1$, $I_2$), we compute $\Delta F$ (explained in the previous section) and sample an initial noise:
\begin{equation}
    x_0 \sim \mathcal{N}(0, I) \enspace.
\end{equation}
The change mask latent is then generated by solving the rectified flow ordinary differential equation (ODE) using Euler integration over equally spaced $T$ steps:
\begin{equation}
    x_{t+\frac{1}{T}} = x_t + \frac{1}{T} \mathcal{M}_\theta([x_t, \Delta F], t) \enspace.
    \label{eq:integrate}
\end{equation}
The final latent $\hat{x} = x_T$ is decoded into a binary RGB change mask using the pretrained VAE decoder $\mathcal{V}^{-1}$:
\begin{equation}
    \hat{M}_{RGB} = \mathcal{V}^{-1}(\hat{x}) \enspace.
\end{equation}
To obtain the final single-channel binary mask, the prediction is averaged across the RGB channels, yielding $\hat{M}$.
This inference process is depicted in Fig.~\ref{fig:arch} (bottom). By using the RF formulation, we allow for a flexible number of time steps at inference, which can be \textit{freely adjusted after training}. 

\subsection{Extension to Semantic Change Detection}
\label{sec:scd}

Semantic change detection (SCD) extends binary CD with a \emph{from--to} transition at each changed location: besides localising change (predicting $M$), it predicts a pre- and post-change semantic map $S_1, S_2 \in \{0,\dots,K\}^{H \times W}$, where $K$ is the number of semantic classes and $0$ is the no-change class~\cite{yang2020second}. We extend our latent generative formulation to SCD by modifying how the semantic change maps are encoded and how the model is conditioned.

\noindent\textbf{Semantic maps as latent images.} To encode semantic change maps to a format appropriate for an RGB VAE, we colourise them with a palette $\mathcal{P}$ that maps each class to a maximally spaced RGB colour, making them easily separable, i.e., $C=\mathcal{P}(S)$. The palette is formed by a greedy max-distance algorithm similar to~\cite{chen2023genSeg} that initialises no-change to white and avoids near-grey codes (details in the Supp.). We find this to perform best, although any well-separated colouring still works (see Supp.). Both $C_1$ and $C_2$ are then encoded by the \emph{same} pretrained VAE as in BCD, and the resulting target representations are obtained \emph{jointly} via channel-concatenation:
\begin{equation}
    y_1 = \big[\,\mathcal{V}(\mathcal{P}(S_1)),\; \mathcal{V}(\mathcal{P}(S_2))\,\big] \enspace,
\end{equation}
allowing the model to capture correlated changes between the two acquisitions rather than segmenting them independently. At inference, a generated latent $\hat y$ is mapped back from RGB to its original class indices by nearest-neighbour assignment to the entries of palette $\mathcal{P}$, i.e., $\hat S = \mathcal{P}^{-1}(\mathcal{V}^{-1}(\hat y))$.

\noindent\textbf{Image-to-semantic rectified flow.} 
Because each semantic change map is itself a labelling of an input image, the image latents are already spatially aligned with the target and provide a far stronger initialisation than noise alone. We therefore utilise channel-concatenated \emph{image pair} VAE latents perturbed by Gaussian noise as starting state:
\begin{equation}
    y_0 = \big[\,\mathcal{V}(I_1),\; \mathcal{V}(I_2)\,\big] + \epsilon, \qquad \epsilon \sim \mathcal{N}(0, I) \enspace,
\end{equation}
and learn a rectified flow that transports $y_0$ to the semantic latents $y_1$. The per-sample noise $\epsilon$ retains the image prior while restoring the stochasticity that drives sampling-based ensembling. Training follows the standard objective:
\begin{equation}
    y_t = (1-t)\,y_0 + t\,y_1, \quad \mathcal{L}_{\text{SCD}} = \big\| (y_1 - y_0) - v_{\text{pred}} \big\|_2^2 \enspace,
\end{equation}
\begin{equation}
    v_{\text{pred}} = \mathcal{M}^{\text{sem}}_\theta \big([\,y_t,\; M,\; F_1,\; F_2,\; \Delta F\,], t\big) \enspace.
\end{equation}
We expand the conditioning from using only $\Delta F$ in the binary case (Eq.~\ref{eq:dlatent}) by channel-concatenating raw bi-temporal features $F_1$ and $F_2$ for semantic guidance and the binary change mask $M$ for binary guidance ($M$ is ground truth at training, binary prediction $\hat M$ at inference obtained with a separate model). Inference is performed in the same way as in the binary case (Fig.~\ref{fig:arch}, Eq.~\ref{eq:integrate}), integrating the field from $y_0$ to produce the final latents $\hat y=y_T$, which are then decoded as specified above. Further details, as well as a flow diagram for SCD, are in the Supp.

\subsection{Sampling-based Ensemble} 
Unlike standard discriminative models~\cite{cheng2025changedino}, ChangeFlow's generative formulation \textit{inherently} enables sampling‑based inference (i.e. ensembling) without additional training. The rectified flow model implicitly defines a conditional distribution~\cite{liu2023rectifiedflow} over change masks by marginalising latent noise, i.e., $p(M \mid \Delta F) = \int p(M \mid \Delta F, x_0)\,p(x_0)\,dx$. In practice, this marginalisation is approximated via Monte Carlo sampling by generating ensemble masks $\hat{M}_{ens}^{(i)}; i\in\{1,\dots,N\}$ starting from different initial noise $x_o^i$ and aggregating them into a joint prediction $\hat{M}$. We use mean aggregation for binary masks and majority-vote aggregation for semantic masks.

\section{Results}
\noindent\textbf{Implementation details.}
\label{par:impldet}
We use DINOv3 ViT-L~\cite{simeoni2025dinov3} as the encoder and extract features from its final layer. For mask encoding, we adopt the VAE from SD-XL~\cite{podell2024sdxl}. To spatially align the encoder and VAE latents, we apply bicubic interpolation. Each inference involves 5 steps (i.e., $T=5$) and an ensemble of 5 predictions. 
Input images are augmented with random flips and rotations during training.
We train using the Muon~\cite{jordan2024muon} optimiser, with an initial learning rate of $10^{-4}$ for DiT and $5\cdot10^{-5}$ for the encoder, and a cosine scheduler without restarts. Training lasts 300 epochs with a batch size of 32 on an NVIDIA A100 GPU. Additional details are in the Supp.

\noindent\textbf{Evaluation metrics and datasets.}
\label{par:metr_data}
We evaluate binary change detection (BCD) performance using \emph{binary} F1, considering only \emph{change class}~\cite{daudt2018fcn, rolih2025btc}, and for semantic change detection (SCD), we use mIoU, SeK and $F_{scd}$, as defined in related work~\cite{ding2024scannet}. Metrics are calculated on the model from the final epoch. For robust evaluation, we benchmark on four BCD datasets that cover \textit{diverse locations}, \textit{sensors}, and \textit{ground sampling distances}, and span \textit{diverse change types}. \textit{SYSU}~\cite{shi2022sysuDSAMnet} covers various change types, from buildings and vegetation to sea changes. \textit{LEVIR}~\cite{chen2020levirStanet} focuses on building changes, while \textit{CLCD}~\cite{li2022clcdMSCANET} captures only changes that happen on croplands. \textit{OSCD}~\cite{daudt2018urban} is a low-resolution global Sentinel-2 dataset covering urban changes. We evaluate SCD on SECOND~\cite{yang2020second}, an established SCD dataset that contains a no-change class and 6 semantic classes. All input images in the BCD setting are cropped to $256 \times 256$ pixels, and in SCD to $512\times512$~\cite{rolih2025btc, yang2020second}. Models are trained on a dedicated training set and evaluated on a held-out test set. Additional details are in the Supp.

\subsection{Main results}

We evaluate ChangeFlow against a range of BCD methods. ChangeDINO~\cite{cheng2025changedino} represents the current state-of-the-art in the BCD setting and uses the same DINOv3~\cite{simeoni2025dinov3} backbone as our ChangeFlow. We summarise quantitative results across all BCD datasets and methods in Tab.~\ref{tab:sota}. Extended results and implementation details are in the Supp.

\begin{table*}[!h]
    \caption{Comparison of our proposed method, ChangeFlow, to the state-of-the-art on four different binary change detection datasets measured with \textbf{F1}. We mark \textcolor{goldD}{first} and \textcolor{silverD}{second} place results. FPS ([img/s]) was benchmarked on an NVIDIA A100 using the protocol described in the Supp. FPS for ChangeFlow includes all computation, including all steps and ensembling.
    }
    \centering
    \centering
    \begin{tabular}{lcc|ccccc}
    \toprule
    	\multirow{2}{*}{~} & FPS & Param. [M]& SYSU& LEVIR& CLCD& OSCD& \textit{Avg}\\
 \hline
ChFormer~\cite{bandara2022changeFormer}\scriptsize{\textcolor{gray}{IGARSS22}} & 36.2 & 41.0 & 77.9 & 89.5 & 60.8 & 48.1 & 69.1\\
SwinSUNet~\cite{zhang2022swinsunet}\scriptsize{\textcolor{gray}{TGRS22}} & 33.1 & 43.6 & 76.6 & 89.3 & 75.8 & 52.8 & 73.6\\
GFM~\cite{mendieta2023gfm}\scriptsize{\textcolor{gray}{CVPR23}} & 44.9 & 120.5 & 81.2 & 89.8 & 77.5 & 54.1 & 75.7\\
GCD-DDPM~\cite{wen2024gcd-ddpm}\scriptsize{\textcolor{gray}{TGRS24}} & 0.02& 131.9 & 64.5 & 80.7 & 46.9 & 7.0 & 49.8\\
BiFA~\cite{zhang2024bifa}\scriptsize{\textcolor{gray}{TGRS24}} & 32.2 & 9.9 & \bm3{83.8} & 89.5 & 74.5 & 37.4 & 71.3\\
MaskCD~\cite{yu2024maskcd}\scriptsize{\textcolor{gray}{TGRS24}} & 6.5 & 107.4 & \bm3{83.8} & 90.3 & 76.6 & 34.7 & 71.4\\
ChMamba~\cite{chen2024changeMamba}\scriptsize{\textcolor{gray}{TGRS24}} & 14.4 & 92.4 & 81.5 & \bm3{91.8} & 80.3 & 45.8 & 74.9\\
MTP~\cite{wang2024mtp}\scriptsize{\textcolor{gray}{JSTARS24}} & 31.2& 107.8 & 81.3 & 91.7 & 80.3 & 52.8 & 76.5\\
HySCDG~\cite{benidir2025hyscdg}\scriptsize{\textcolor{gray}{CVPR25}} & 41.0 & 65.1 & 78.7 & 91.1 & 64.3 & 53.6 & 71.9\\
DDPM-CD~\cite{bandara2025ddpmcd}\scriptsize{\textcolor{gray}{WACV25}} & 4.6 & 437.5 & 80.5 & 90.9 & 71.4 & 37.1 & 70.0\\
SatDiFuser~\cite{jia2025satdifuser}\scriptsize{\textcolor{gray}{ICCV25}} & 1.8 & 1413.6 & 82.0 & 90.2 & 79.1 & \bm3{55.2} & 76.6\\
BTC~\cite{rolih2025btc}\scriptsize{\textcolor{gray}{TGRS25}} & 32.4 & 120.1 & 82.4 & 91.5 & \bm3{80.9} & 54.3 & \bm3{77.3}\\
ChangeDINO (DINOv3)~\cite{cheng2025changedino}\scriptsize{\textcolor{gray}{arXiv25}} & 8.9 & 311.1 & \bm2{83.9} & \bm1{92.2} & \bm2{81.4} & \bm2{58.8} & \bm2{79.1}\\
\textbf{ChangeFlow} & 11.8 & 403.3 & \bm1{85.6} & \bm2{92.1} & \bm1{84.5} & \bm1{59.5} & \bm1{80.4}\\
     \bottomrule
    \end{tabular}
    \label{tab:sota}
\end{table*}

\noindent \textbf{Quantitative results.} ChangeFlow achieves the best average F1 of $80.4\%$, $1.3$ points above the previous best, ChangeDINO, and sets a new state of the art on SYSU, CLCD, and OSCD ($85.6\%$, $84.5\%$, and $59.5\%$). On LEVIR, it remains within $0.1$ points of the best competing method. At $11.8$ images per second, ChangeFlow is faster than ChangeDINO by roughly $3$ images per second, while every method with higher throughput trails it by $3.1$ to $11.3$ F1 percentage points (p.p.). 

\noindent\textbf{Comparison to diffusion-based methods.} ChangeFlow outperforms all prior discriminative approaches that use diffusion models as feature extractors. In particular, it outperforms DDPM-CD by 10.3 p.p., and the recent SatDiFuser foundation model by 4.8 p.p. 
It also substantially exceeds the pixel-space generative baseline, GCD-DDPM, by 30.5 p.p. and is almost 3 orders of magnitude faster at inference, demonstrating our substantial gains to both performance and speed compared to the previous generative attempts.

\begin{figure}[!t]
    \centering
    \includegraphics[width=1\linewidth]{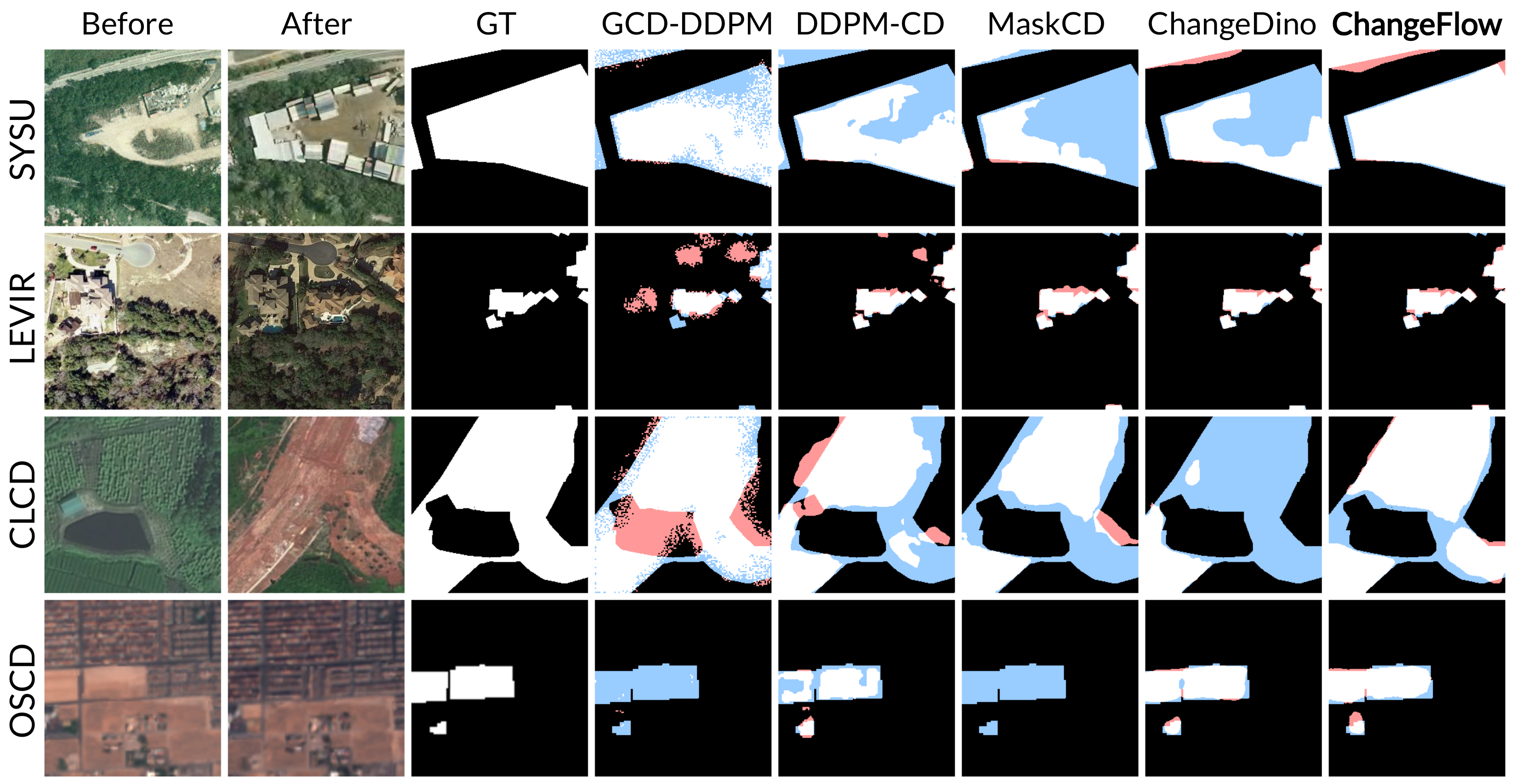}
    \caption{Qualitative comparison of competing methods. The pair of images under consideration is shown in the first and second columns, followed by the ground-truth mask and the predictions from related methods and our method. \textcolor{falsePos}{False positives are marked in red} and \textcolor{falseNeg}{false negatives in blue}.}
    \label{fig:qual}
\end{figure}

\noindent \textbf{Qualitative results.} In Fig.~\ref{fig:qual}, we show a qualitative comparison between evaluated methods. Compared with ChangeDINO, our method reduces missed detections in homogeneous regions, consistent with its coherent mask-generation behaviour. Compared to DDPM-CD, which uses diffusion primarily as a feature extractor, ChangeFlow better recovers complete change regions and reduces both false positives and false negatives. Even MaskCD's~\cite {yu2024maskcd} instance-based predictions remain fragmented across diverse change types, whereas ChangeFlow directly generates a globally consistent mask. More qualitative results, including failure cases, are in the Supp.

\begin{figure}[!t]
    \centering
    \includegraphics[width=1\linewidth]{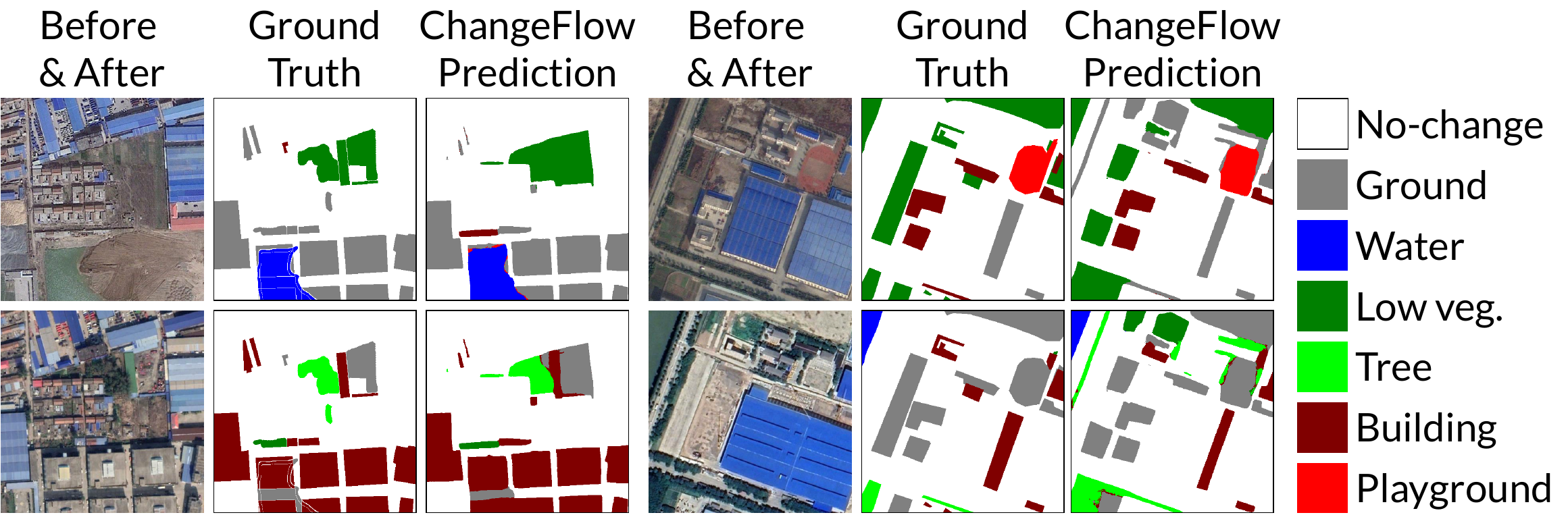}
    \caption{Qualitative comparison of SCD on SECOND. The pair of images under consideration is shown in the first row, followed by the ground-truth mask and the prediction.}
    \label{fig:scd}
\end{figure}

\begin{table}[!h]
    \caption{SCD results evaluated on the SECOND dataset.}
    \setlength{\tabcolsep}{2pt}
    \centering
    \begin{tabular}{lccc}
    \toprule
     & mIoU & SeK & $\text{F}_\text{scd}$\\
\midrule

SCanNet~\cite{ding2024scannet}\scriptsize{\textcolor{gray}{TGRS24}} & 73.4 & 23.6& 63.8 \\
LSAFNet~\cite{zhou2024lsafnet}\scriptsize{\textcolor{gray}{GRSL24}} & 73.7& 24.3& 64.5 \\
ChMamba~\cite{chen2024changeMamba}\scriptsize{\textcolor{gray}{TGRS24}} & 73.5& 23.9 & 64.0 \\
Change3D~\cite{zhu2025change3d}\scriptsize{\textcolor{gray}{CVPR25}} & 73.0 & 23.0 & 62.8 \\
UniChange~\cite{zhang2026unichange}\scriptsize{\textcolor{gray}{CVPR26}} & 72.9 & 23.0 & 63.5 \\
TaCo~\cite{guo2025taco}\scriptsize{\textcolor{gray}{arXiv25}} & \bm2{73.8} & \bm2{24.7} & \bm2{64.9} \\
\textbf{ChangeFlow} & \bm1{73.9} & \bm1{25.3} & \bm1{65.9}\\
         \bottomrule
    \end{tabular}
    \label{tab:scd}
\end{table}
\noindent \textbf{SCD results.} We compare ChangeFlow against a range of recent SCD methods on SECOND~\cite{yang2020second} and present results in Tab.~\ref{tab:scd}. ChangeFlow achieves a new state-of-the-art with a $\text{F}_\text{scd}$ of 65.9, 1 p.p. higher than the previous best TaCo~\cite{guo2025taco}. Similar to the binary setting, our formulation leads to good coherence within change regions and yields strong results, as is also evident in the qualitative results shown in Fig.~\ref{fig:scd}.

\begin{figure}[!h]
    \centering
    \includegraphics[width=1\linewidth]{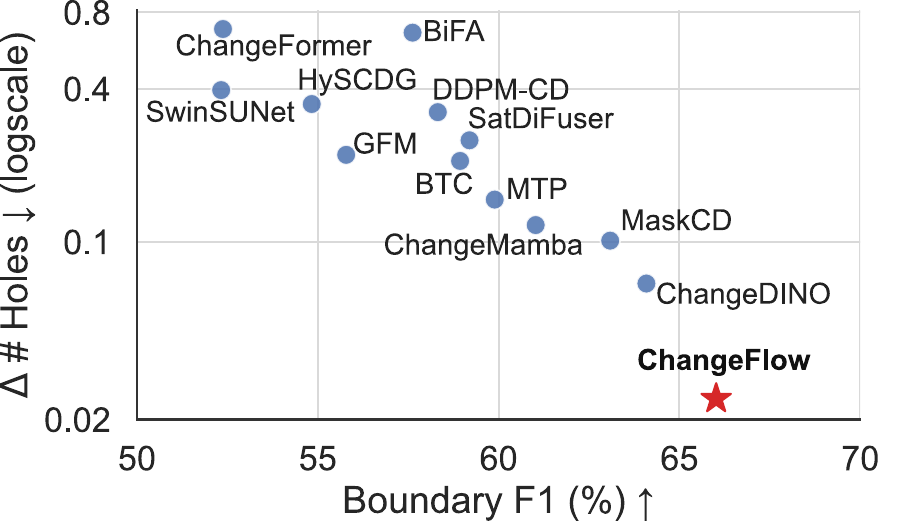}
    \caption{Coherence measured as hole count error ($\Delta$ \#Holes, lower is better) and border F1 (higher is better) averaged over 4 BCD datasets.}
    \label{fig:coh_all}
\end{figure}

\noindent \textbf{Coherence analysis.} ChangeFlow's generative formulation enables inherent global prediction coherence. 
To quantitatively evaluate this, we assess structural consistency by calculating the error relative to the expected ground-truth number of holes (reported as $\Delta$ \#Holes) and boundary F1 across 4 BCD datasets. Fig.~\ref{fig:coh_all} shows that ChangeFlow yields low structural error, indicating the fewest spurious holes and highest accuracy in border regions. Details are in the Supp.

\subsection{Ablation study}

We isolate the impact of our contributions by ablating key design choices. Implementation details and additional ablations are in the Supp. (e.g. SCD ablations).

\begin{table}[!t]
    \caption{Ablation of core design choices measured with \textbf{F1}.}
    \setlength{\tabcolsep}{2pt}
    \centering
    \begin{tabular}{lccccc}
    \toprule
     & SYSU& LEVIR& CLCD& OSCD& \textit{Avg}\\
\midrule
Discriminative CF & 84.4 & 92.1 & 83.6 & 57.2 & 79.3\\
No ensem., gen. CF  & 84.2& 91.9& 84.5& 58.8& 79.8 \\
VAE finetune & 81.2 & 92.0 & 81.9 & 36.3 & 72.9\\
\midrule
Complex cond. & 85.1 & 92.1 & 83.9 & 57.3 & 79.6\\
No norm. cond. & 81.8 & 91.4 & 77.6 & 56.4 & 76.8\\
No abs. cond. & 81.2 & 91.7 & 82.2 & 35.6 & 72.7\\
\midrule
 \rowcolor{gray!20} \textit{Ours} &  85.6& 92.1& 84.5& 59.5& 80.4\\ 
         \bottomrule
    \end{tabular}
    \label{tab:abl_arch}
\end{table}

\noindent\textbf{Core ablations.} Tab.~\ref{tab:abl_arch} (top block) isolates the generative formulation itself. Compared to our main model in Tab.~\ref{tab:sota}, \textit{Discriminative CF} keeps the architecture and parameter count identical, but replaces the rectified-flow objective (Eq.~\ref{eq:rf_loss}) with a discriminative Dice loss~\cite{Milletari2016diceVnet} on the decoded mask, with gradients being passed through the frozen VAE, while also using a single sample during inference. Against this strong, equally equipped discriminative baseline, our advantage is twofold. \textbf{(i)~The generative objective.} At a single prediction (\textit{No ensem., gen. CF}), compared to the discriminative baseline, the generative objective alone raises average F1 by $0.5$ points ($79.3\!\rightarrow\!79.8$), while \textit{also exceeding} the previous state of the art, ChangeDINO, by $0.7$ points (Tab.~\ref{tab:sota}, $79.1\!\rightarrow\!79.8$).
\textbf{(ii)~The inherent ensembling.} Aggregating five samples (\textit{Ours}), a capability our generative formulation inherently enables, adds a further $0.6$ points ($79.8\!\rightarrow\!80.4$). Ensembling thus strengthens our results but is not the sole source of performance gains: the generative formulation (\textit{No ensem., gen. CF}) is state-of-the-art on its own, and the two effects are additive. Finally, fine-tuning the VAE (\textit{VAE finetune}) degrades performance, consistent with prior findings~\cite{bagchi2025refereverything}, so we keep it frozen.

\noindent\textbf{Conditioning ablations.} The second block of Tab.~\ref{tab:abl_arch} isolates the conditioning signal. Replacing our parameter-free feature difference (Eq.~\ref{eq:dlatent}) with a learnable convolutional fusion module (\textit{Complex cond.}) lowers average F1, confirming that elaborate conditioning is counterproductive and our simpler signal is more effective. Ablating its two components shows that both are essential: removing normalisation (\textit{No norm.\ cond.}) and removing the absolute value (\textit{No abs.\ cond.}) degrade performance, underscoring the necessity of an order- and magnitude-invariant signal.

\begin{figure}[!t]
    \centering
    \includegraphics[width=1\linewidth]{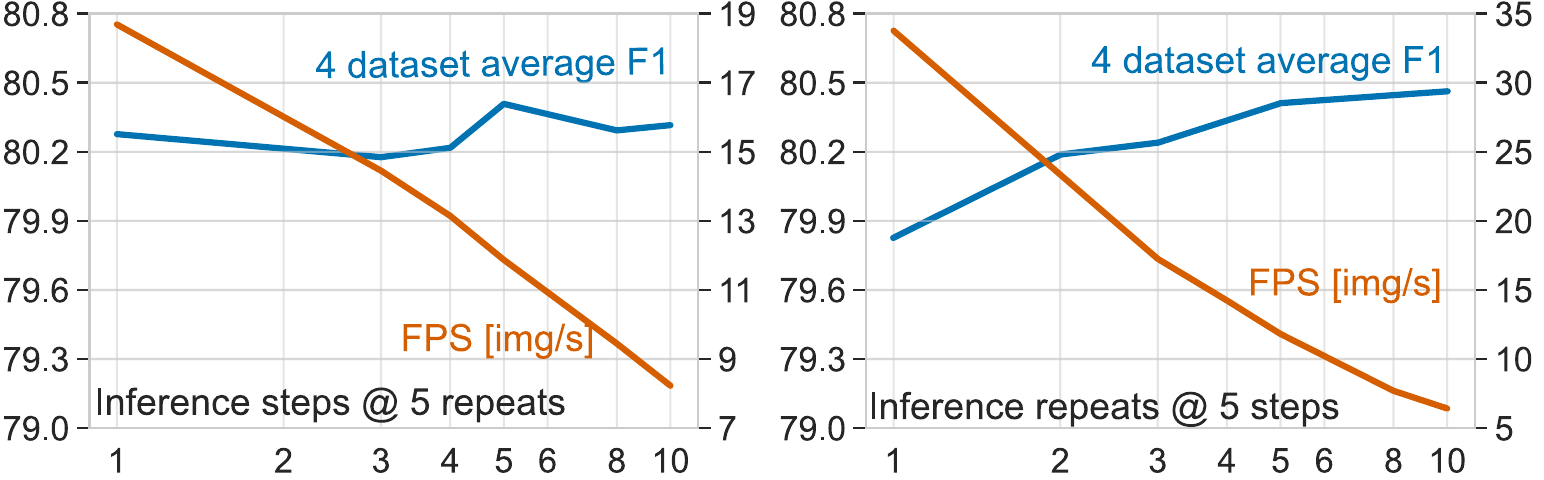}
    \caption{Impact of number of steps and inference repetitions (ensemble size). Change detection performance is reported on the left y-axis (average F1 across 4 BCD datasets), while inference speed is reported on the right y-axis (frames per second - FPS; protocol in the Supp.).}
    \label{fig:step_rep}
\end{figure}

\noindent\textbf{Inference steps and ensembling analysis.}
ChangeFlow uses $T=5$ steps and 5 predictions in the ensemble (repetitions). Fig.~\ref{fig:step_rep} shows that increasing repetitions at fixed steps yields more gains than increasing $T$ beyond a small number of steps at fixed repetitions, while both increase runtime. This provides a controllable speed--accuracy trade-off at inference time.

\noindent\textbf{Limitations and future work.} The main limitation of our formulation is computational: inference cost scales with the number of ensembled samples. However, even a single sample already exceeds the previous state of the art (Tab.~\ref{tab:abl_arch}), making additional samples an optional boost in accuracy. We also deliberately build on an off-the-shelf pretrained VAE; since information loss is minimal (Tab.~\ref{tab:vae_pass}), a mask-specialised VAE is unnecessary, though it remains a promising future refinement. More broadly, casting change detection as latent generation opens interesting directions, notably the use of textual guidance for open-vocabulary change detection.

\section{Conclusion}

We introduced ChangeFlow, a latent generative framework that recasts remote sensing change detection as the synthesis of a change mask rather than per-pixel classification. By encoding masks with a pretrained VAE and generating them with a rectified flow, ChangeFlow models the mask as a single, coherent object while avoiding the pixel-space cost and elaborate conditioning of prior generative approaches. This yields properties that discriminative pipelines obtain only through additional machinery: spatially coherent predictions; a controllable speed--accuracy trade-off at inference; and sampling-based ensembling at no extra training cost. These structural benefits are demonstrated quantitatively: among all evaluated methods, ChangeFlow produces masks with the fewest spurious holes. Across four binary benchmarks, ChangeFlow reaches an average F1 of $80.4\%$, improving on the previous best by $1.3$ points and setting a new state of the art on SYSU, CLCD, and OSCD, while remaining competitive on LEVIR. The formulation extends to semantic change detection, achieving a new state-of-the-art $65.9$ $\text{F}_\text{scd}$ on SECOND. More broadly, our results indicate that latent generative inference is an efficient and conceptually distinct alternative to discriminative dense prediction, and point toward flow-based mask synthesis as a promising direction for other dense prediction tasks.

\section*{Acknowledgements}
This work was in part supported by the ARIS research projects GC-0006 (GeoAI) and J2-60045 (RoDEO), research programme P2-0214, and the supercomputing network SLING (ARNES, EuroHPC Vega).

\bibliography{main}

\maketitlesupplementary
 
\appendix
\crefalias{section}{appendix}
\crefalias{subsection}{appendix}
\setcounter{secnumdepth}{2}

In these supplementary materials, we provide additional details that extend beyond the scope of the main manuscript. It is organised as follows: 
\begin{itemize}
    \item \textbf{Extended dataset details} in Section~\ref{a:data}.
    \item \textbf{Extended results} with additional metrics (precision and recall), and additional BCD and SCD ablations in Section~\ref{a:ext_res}.
    \item \textbf{Additional qualitative results}, including failure cases, more SCD results, VAE mask reconstruction, and intermediate step generation in Section~\ref{a:qual}.
    \item \textbf{Computational efficiency} protocol, extended results, and discussion in Section~\ref{a:comp}.
    \item \textbf{Extended implementation details} for model and training in BCD and SCD setting (including a flow diagram for SCD case), our ablations and analyses, and related methods in Section~\ref{a:ext_impl}.
\end{itemize}

\section{Extended dataset details}
\label{a:data}

\begin{table*}[!h]
\resizebox{\linewidth}{!}{
\setlength{\tabcolsep}{3pt}
    \begin{tabular}{lccccccccc}
        \toprule
 & Acquisition & Resolution & Change Type & Interval & Region & \makecell{Image count \\ train\textbackslash val\textbackslash test} & Patch & \makecell{Changed \\ Pixels} & \makecell{Unchanged \\ Pixels} \\
\midrule
SYSU (Shi et al.) & Aerial & 0.5m & \makecell{Building, urban,\\groundwork, road,\\vegetation, sea} & 2007-2014 & Hong Kong & \makecell{12000\\4000\\4000} & $256\times256$ & $21.8~\%$ & $78.2~\%$ \\\midrule
LEVIR~\cite{chen2020levirStanet} & \makecell{Google Earth \\ satellite} & 0.5m & Building & 2002-2018 & \makecell{20 regions \\ in US} & \makecell{7120\\1024\\2048} & $256\times256$ & $4.7~\%$ & $95.3~\%$ \\\midrule
CLCD~\cite{li2022clcdMSCANET} & \makecell{Satellite \\ (Gaofen-2)} & 0.5m-2m & \makecell{Multiple types\\limited to\\croplands} & 2017-2019 & \makecell{Guangdong,\\ China} & \makecell{1440\\480\\480} & $256\times256$ & $7.6~\%$ & $92.4~\%$ \\\midrule
OSCD~\cite{daudt2018urban} & \makecell{Satellite \\ (Sentinel-2)} & 10m & Urban & 2015-2018 & \makecell{24 regions \\ worldwide} & \makecell{827\\-\\385} & $96\times96$ & $3.2~\%$ & $96.8~\%$ \\
\midrule
SECOND~\cite{yang2020second} & \makecell{Aerial \\ (multiple sensors)} & undisclosed & \makecell{Semantic dataset:\\ground, tree, low vegetation\\ water, buildings\\ playgrounds} & undisclosed & \makecell{Hangzhou, Chengdu,\\Shanghai, China} & \makecell{2968\\-\\1694} & $512\times512$ & 19.1 \% & 80.9 \% \\
\bottomrule
    \end{tabular}
    }
    \caption{Additional details for the datasets used in the paper.}
    \label{atab:data_extend}
\end{table*}

Additional dataset details are provided in \Cref{atab:data_extend}. Our benchmarks span diverse change scenarios, including building and urban expansion, as well as changes limited to croplands. They also vary substantially in ground sampling distance (GSD), acquisition sensor, and scale, ranging from a few hundred to several thousand image pairs. This diversity strengthens the robustness and generality of our conclusions. 

A recurring challenge in RSCD is severe class imbalance: changed pixels typically constitute less than 10\% of all pixels. SYSU and SECOND are exceptions, exhibiting a higher (but still relatively unbalanced) change ratio.

\subsection{Data implementation details}
\label{asub:data_impl}

\textbf{Dataset splits.} Official train and test splits are used for OSCD, SYSU, CLCD, LEVIR, and SECOND to ensure full reproducibility and fair comparison. We also use a validation set (for optimal threshold computation) from SYSU, CLCD, and LEVIR.

The HuggingFace public source for the data used is as follows:
\begin{itemize}
    \item SYSU: \textcolor{weights}{ericyu: SYSU\_CD}
    \item LEVIR: \textcolor{weights}{ericyu: LEVIRCD\_Cropped256}
    \item CLCD: \textcolor{weights}{ericyu: CLCD\_Cropped\_256}
    \item OSCD: \textcolor{weights}{blaz-r: OSCD\_RGB\_Cropped\_96}
\end{itemize}
SECOND was obtained via GitHub \textcolor{weights}{captain-whu: SCD}.

Data pre-processing details are listed in \Cref{a:ext_impl}.

\section{Extended results}
\label{a:ext_res}

This section provides extended results (with additional metrics) in \Cref{a:ext_main}, additional BCD ablations in \Cref{a:ext_abl}, and additional SCD ablation in \Cref{asub:scd_abl}.

\subsection{Main results with additional metrics}
\label{a:ext_main}

\Cref{atab:main} presents the results from the main paper with additional precision and recall metrics. In addition to the discussion in the main body of the paper, we note that ChangeFlow consistently achieves high recall while maintaining a balance between precision and recall. Compared to the previous best overall method, ChangeDINO, our method achieves a recall that is almost 5 percentage points higher. It does, however, achieve lower precision, but when these two are combined in F1, our method achieves a better balance.

\begin{table*}[!ht]
    \centering
    \caption{Change detection results across four diverse datasets and their average. We report Precision (Pr.), Recall (Re.), and F1 score. \textcolor{goldD}{First}, \textcolor{silverD}{second}, and \textcolor{bronzeD}{third} place results are marked.}
    \label{atab:main}
    \setlength{\tabcolsep}{2pt}
     \resizebox{1\linewidth}{!}{
    \begin{tabular}{lccc|ccc|ccc|ccc|ccc}
    \toprule
    & \multicolumn{3}{c}{SYSU} & \multicolumn{3}{c}{LEVIR} & \multicolumn{3}{c}{CLCD} & \multicolumn{3}{c}{OSCD} & \multicolumn{3}{c}{\textit{Avg}} \\
 &Pr.& Re. & F1 &Pr.& Re. & F1 &Pr.& Re. & F1 &Pr.& Re. & F1 &Pr.& Re. & F1\\
 \hline
FC-Siam-Diff~\cite{daudt2018fcn}\tiny{{ICIP18}} & 83.5 & 61.5 & 70.8 & 83.0 & 80.6 & 81.8 & 54.0 & 54.3 & 54.1 & 27.8 & \bm1{68.5} & 39.4 & 62.1 & 66.2 & 61.5\\
ChFormer~\cite{bandara2022changeFormer}\tiny{{IGARSS22}} & 82.8 & 73.5 & 77.9 & 91.7 & 87.3 & 89.5 & 61.4 & 60.4 & 60.8 & 60.2 & 40.1 & 48.1 & 74.0 & 65.3 & 69.1\\
SwinSUNet~\cite{zhang2022swinsunet}\tiny{{TGRS22}} & 89.2 & 67.2 & 76.6 & 86.9 & \bm2{91.7} & 89.3 & 79.5 & 72.5 & 75.8 & 61.7 & 46.3 & 52.8 & 79.3 & 69.4 & 73.6\\
GFM~\cite{mendieta2023gfm}\tiny{{CVPR23}} & \bm2{89.7} & 74.3 & 81.2 & 90.8 & 88.8 & 89.8 & 82.2 & 73.2 & 77.5 & 55.9 & 52.5 & 54.1 & 79.6 & 72.2 & 75.7\\
GCD-DDPM~\cite{wen2024gcd-ddpm}\tiny{{TGRS24}} & 54.5 & 78.9 & 64.5 & 79.0 & 82.6 & 80.7 & 42.4 & 52.3 & 46.9 & 48.3 & 3.7 & 7.0 & 56.1 & 54.4 & 49.8\\
BiFA~\cite{zhang2024bifa}\tiny{{TGRS24}} & 87.4 & \bm2{80.4} & \bm3{83.8} & 90.9 & 88.1 & 89.5 & 79.4 & 70.1 & 74.5 & 61.5 & 27.1 & 37.4 & 79.8 & 66.4 & 71.3\\
MaskCD~\cite{yu2024maskcd}\tiny{TGRS24} & 88.0 & 80.0 & \bm3{83.8} & 91.5 & 89.2 & 90.3 & 79.5 & 73.9 & 76.6 & 60.9 & 24.3 & 34.7 & 80.0 & 66.8 & 71.4\\
ChMamba~\cite{chen2024changeMamba}\tiny{{TGRS24}} & \bm3{89.6} & 74.7 & 81.5 & 92.4 & 91.2 & \bm3{91.8} & \bm1{87.3} & 74.4 & 80.3 & 63.4 & 36.1 & 45.8 & 83.2 & 69.1 & 74.9\\
MTP~\cite{wang2024mtp}\tiny{{JSTARS24}} & 88.5 & 75.2 & 81.3 & \bm3{92.8} & 90.7 & 91.7 & 85.4 & 75.8 & 80.3 & 43.9 & \bm2{66.5} & 52.8 & 77.6 & \bm2{77.0} & 76.5\\
HySCDG~\cite{benidir2025hyscdg}\tiny{CVPR25} & 83.3 & 74.6 & 78.7 & 92.6 & 89.7 & 91.1 & 71.1 & 58.8 & 64.3 & \bm3{64.4} & 45.9 & 53.6 & 77.8 & 67.2 & 71.9\\
DDPM-CD~\cite{bandara2025ddpmcd}\tiny{{WACV25}} & 87.3 & 74.7 & 80.5 & \bm1{93.1} & 88.8 & 90.9 & 78.9 & 65.2 & 71.4 & 61.9 & 26.5 & 37.1 & 80.3 & 63.8 & 70.0\\
SatDiFuser~\cite{jia2025satdifuser}\tiny{ICCV25} & 88.6 & 76.3 & 82.0 & 91.0 & 89.3 & 90.2 & 86.2 & 73.0 & 79.1 & \bm2{72.9} & 44.4 & \bm3{55.2} & \bm2{84.7} & 70.8 & 76.6\\
BTC~\cite{rolih2025btc}\tiny{{TGRS25}} & \bm1{90.2} & 75.8 & 82.4 & 92.7 & 90.3 & 91.5 & 86.2 & \bm3{76.2} & \bm3{80.9} & 64.1 & 47.1 & 54.3 & \bm3{83.3} & 72.3 & \bm3{77.3}\\
ChangeDINO~\cite{cheng2025changedino}\tiny{arXiv25} (DINOv3) & 88.2 & \bm3{80.1} & \bm2{83.9} & \bm2{92.9} & \bm3{91.4} & \bm1{92.2} & \bm2{86.6} & \bm2{76.7} & \bm2{81.4} & \bm1{73.0} & 49.3 & \bm2{58.8} & \bm1{85.2} & \bm3{74.4} & \bm2{79.1}\\
\textbf{ChangeFlow}\tiny{(5step, 5rep)} & 86.8 & \bm1{84.3} & \bm1{85.6} & 91.5 & \bm1{92.7} & \bm2{92.1} & \bm3{86.3} & \bm1{82.8} & \bm1{84.5} & 62.2 & \bm3{57.0} & \bm1{59.5} & 81.7 & \bm1{79.2} & \bm1{80.4}\\
\hline
ChangeFlow \tiny{(1step, 5rep)} &86.7& 84.5& 85.6& 91.4& 92.6& 92.0& 86.5& 82.6& 84.5& 65.5& 53.7& 59.0& 82.5& 78.4& 80.3\\
ChangeFlow \tiny{(5step, 1rep)}&87.5& 81.3& 84.2& 91.5& 92.4& 91.9& 87.0& 82.2& 84.5& 66.3& 52.8& 58.8& 82.9& 77.1& 79.8\\
    \bottomrule
    \end{tabular}
    }
\end{table*}

\subsection{Additional ablations}
\label{a:ext_abl}

In this subsection, we present additional ablations. First, we report an additional experiment with a training time-step-sampling alternative. Next, we present approaches to conditioning vector resizing beyond bicubic interpolation and evaluate normalisation layers beyond LayerNorm. We also evaluate two other simple options when constructing conditioning. We then present encoder and VAE ablations and, finally, study the optimal binarisation strategy in BCD. For visual results (including VAE mask reconstructions), refer to \Cref{a:qual}. Implementation details are in \Cref{asub:abl_imp}. SCD ablations are in \Cref{asub:scd_abl}.

\subsubsection{Training time-step sampling approach.}
\label{asub:time_abl}

\begin{figure}
    \centering
    \includegraphics[width=1\linewidth]{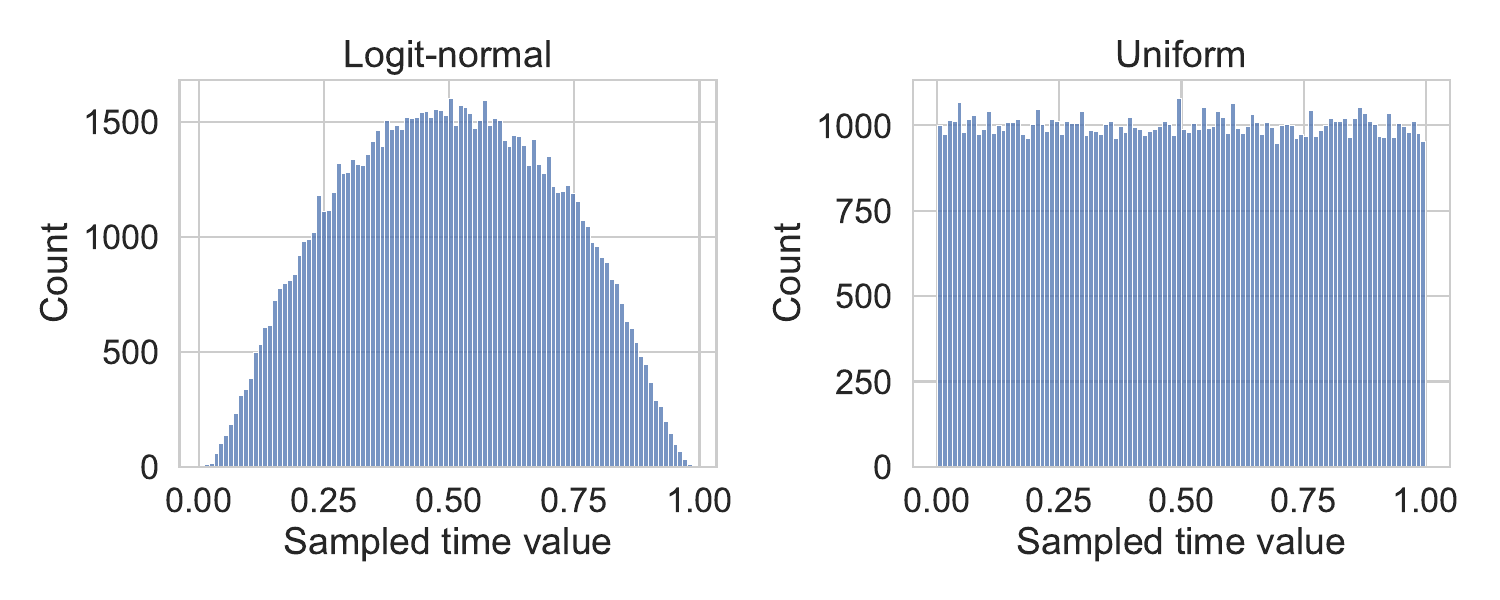}
    \caption{Histogram of 100,000 sampled timesteps in logit-normal and uniform fashion. ChangeFlow uses logit-normal sampling, which emphasises learning at the critical halfway point between noise and data.}
    \label{afig:time_samp}
\end{figure}

ChangeFlow uses logit-normal time sampling during training~\cite{esser2024sd}. This type of sampling emphasises time-steps around 0.5, which is a halfway point between noise and data. This is achieved by sampling from the normal distribution $\mathcal{N}(0, 1)$ and applying a sigmoid to the value, which maps the time to the interval $[0, 1]$. The resulting sampled time is thus concentrated around 0.5, focusing training on the most critical point on the straight line, where paths are most likely to cross and require the most rectification~\cite{liu2023rectifiedflow}. 
A commonly used alternative is uniform sampling on the interval $[0, 1]$, which assigns equal probability to all points. For easier visualisation, a histogram of 100,000 sampled steps in both manners is presented in \Cref{afig:time_samp}

To demonstrate that logit-normal sampling is important for ChangeFlow, we also evaluate a uniform alternative and present the results in \Cref{atab:ln}. Uniform sampling consistently performs worse across all datasets, underscoring the importance of focusing training on the more critical halfway point in the rectified flow.

\begin{table*}[!ht]
    \centering
        \caption{Training time-step sampling ablation results.}
    \label{atab:ln}
    \begin{tabular}{lccccc}
    \toprule
    	\multirow{2}{*}{~}& SYSU& LEVIR& CLCD& OSCD& \textit{Avg.}\\
& F1& F1& F1& F1& F1\\
 \hline
 \rowcolor{gray!20}Logit-normal t sampling (\textit{Ours}) &  85.6& 92.1& 84.5& 59.5& 80.4\\
 Uniform t sampling & 85.5 & 91.5 & 83.5 & 57.8 & 79.6\\
    \bottomrule
    \end{tabular}
\end{table*}

\subsubsection{Conditioning resizing.}
\label{asub:cond_res}

Since the spatial dimension of the VAE latent space may not align with that of the image encoder, some form of resizing is required. In our case, the height and width dimensions of the DINOv3 encoder latent are half the size of the VAE's latent (downsampling by 16 vs 8). To match the dimensions, we use bicubic interpolation to rescale the conditioning vector (which comes from features from the image encoder). We also explored some alternatives, with results presented in \Cref{atab:resize}. A future possibility would also be some form of learnable upscaling. Current results indicate that bicubic achieves the best overall performance, while bilinear outperforms it on SYSU. Lanczos performs worst overall, but all 3 approaches are relatively similar, indicating that this choice is important but not to the extent of other architectural decisions, such as normalisation layers.

\begin{table}[!ht]
    \centering
        \caption{Conditioning vector resizing approach ablation.}
    \setlength{\tabcolsep}{4pt}
    \label{atab:resize}
    \begin{tabular}{lccccc}
    \toprule
    	\multirow{2}{*}{~}& SYSU& LEVIR& CLCD& OSCD& \textit{Avg.}\\
& F1& F1& F1& F1& F1\\
 \hline
 \rowcolor{gray!20}Bicubic (\textit{Ours}) &  85.6& 92.1& 84.5& 59.5& 80.4\\
 Bilinear & 85.9 & 92.1 & 83.9 & 57.8 & 79.9\\
Lanczos & 84.5 & 92.1 & 83.8 & 58.8 & 79.8\\

    \bottomrule
    \end{tabular}
\end{table}

\subsubsection{Different normalisation layers.}
\label{asub:norm_ab}

ChangeFlow uses LayerNorm~\cite{ba2016layer} for conditioning feature normalisation, a common normalisation layer in recent architectures. We also evaluated two other options: InstanceNorm and BatchNorm. Results are presented in \Cref{atab:norm}. LayerNorm achieves the best overall performance, while both alternatives perform considerably worse. This can be explained by the general properties of normalisation layers: BatchNorm depends on batch statistics and can introduce instability when the batch contains heterogeneous bi-temporal pairs, whereas InstanceNorm removes instance-specific contrast information useful for change detection. In contrast, LayerNorm normalises features along the channel dimension of each spatial location independently of other samples, preserving per-pixel structure while ensuring consistent feature scaling. These properties make LayerNorm particularly well-suited for conditioning generative models, yielding the strongest performance in our setting.

\begin{table}[!ht]
    \centering
        \caption{Normalisation layer ablation.}
    \label{atab:norm}
    \setlength{\tabcolsep}{2pt}
    \begin{tabular}{lccccc}
    \toprule
    	\multirow{2}{*}{~}& SYSU& LEVIR& CLCD& OSCD& \textit{Avg.}\\
& F1& F1& F1& F1& F1\\
 \hline
 \rowcolor{gray!20}LayerNorm (\textit{Ours}) &  85.6& 92.1& 84.5& 59.5& 80.4\\
 InstanceNorm & 84.0 & 91.6 & 80.8 & 56.6 & 78.3\\
BatchNorm & 84.4 & 92.0 & 83.7 & 56.2 & 79.1\\
L2 Norm & 85.2 & 92.0 & 84.2 & 57.2 & 79.6\\

    \bottomrule
    \end{tabular}
\end{table}

\subsubsection{Conditioning ablations}

We already present complex conditioning (learnable with a conv layer) in the main paper and also include it here. In \Cref{atab:cond_ab}, we also demonstrate that subtraction, or concatenation, is inferior to the absolute difference that we use. The concatenation option makes the task of finding differences much harder, while the option without absolute value makes the conditioning sensitive to temporal order. We also include the no-normalisation option here, with other normalisation layers in the ablation subsection above.

\begin{table}[!h]
    \caption{Ablation of design choices when constructing conditioning.}
    \setlength{\tabcolsep}{4pt}
    \centering
    \begin{tabular}{lccccc}
    \toprule
     & SYSU& LEVIR& CLCD& OSCD& \textit{Avg}\\
 & F1& F1& F1& F1 & F1\\
\midrule
 \rowcolor{gray!20} SubAbs (Ours)&  85.6& 92.1& 84.5& 59.5& 80.4\\
 Complex & 85.1 & 92.1 & 83.9 & 57.3 & 79.6\\
 No abs. & 81.2 & 91.7 & 82.2 & 35.6 & 72.7\\
 Concat& 77.8 & 91.4 & 80.9 & 21.9 & 68.0\\
 No norm. & 81.8 & 91.4 & 77.6 & 56.4 & 76.8\\
         \bottomrule
    \end{tabular}
    \label{atab:cond_ab}
\end{table}

\subsubsection{Encoder ablations.}
In Table~\ref{tab:abl_enc}, we show that DINOv3 on average provides the strongest features for ChangeFlow. The satellite-pretrained variant (DINOv3 Sat.) and DINOv2 perform worse, most likely due to reduced generalisation from smaller pretraining corpora. RADIO yields solid results, with version 4 performing considerably better than 2.5, but does not surpass plain DINOv3.

\begin{table}[!h]
    \caption{Ablation of shared weight encoder selection.}
    \setlength{\tabcolsep}{2pt}
    \centering
    \begin{tabular}{lccccc}
    \toprule
    & SYSU& LEVIR& CLCD& OSCD& \textit{Avg}\\
    & F1& F1& F1& F1& F1\\
\midrule
\rowcolor{gray!20}DINOv3 &  85.6& 92.1& 84.5& 59.5& 80.4\\
DINOv3 Sat. & 83.6 & 91.6 & 80.7 & 59.4 & 78.9\\
DINOv2 & 78.4 & 91.6 & 78.4 & 54.8 & 75.8\\
RADIO 2.5 & 80.7 & 91.3 & 78.7 & 58.8 & 77.4\\
RADIO 4 & 84.2 & 91.9 & 82.8 & 57.8 & 79.2\\
         \bottomrule
    \end{tabular}
    \label{tab:abl_enc}
\end{table}

\begin{table}[h]
    \caption{Ablation of VAEs used for encoding labels during training and decoding of binary masks during inference.}
    \setlength{\tabcolsep}{2pt}
    \centering
    \begin{tabular}{lcccccc}
    \toprule
     & SYSU& LEVIR& CLCD& OSCD& \textit{Avg}\\
     & F1& F1& F1& F1& F1\\
\midrule
\rowcolor{gray!20}SD-XL VAE &  85.6& 92.1& 84.5& 59.5& 80.4\\
SD 3.5 & 84.4 & 91.6 & 84.4 & 56.7 & 79.3\\
Z-Image & 85.2 & 91.7 & 83.2 & 57.8 & 79.5\\
Flux.1-dev VAE & 84.7 & 91.7 & 82.1 & 57.2 & 78.9\\
CNN Decoder & 84.1 & 89.4 & 83.4 & 54.7 & 77.9\\
         \bottomrule
    \end{tabular}
    \label{tab:abl_vae}
\end{table}

\subsubsection{VAE ablations.}
In Table~\ref{tab:abl_vae}, we compares VAEs for mask encoding and decoding. Among pretrained VAEs, the SD-XL VAE~\cite{podell2024sdxl}, with a latent dimension of 4, achieves the best average performance. In contrast, the VAEs with latent dimension of 16 (SD~3.5~\cite{esser2024sd}, Z-image~\cite{cai2025zimg}, and Flux. 1-dev~\cite{flux2024}) are consistently slightly worse. A plausible explanation is that the higher latent dimensionality increases the difficulty of learning a well-conditioned rectified flow transport for sparse binary masks. Importantly, all pretrained VAEs remain competitive overall, suggesting that off-the-shelf VAE latents provide practical and effective representations for change-mask generation.

Finally, we replace the VAE decoder with a lightweight UNet-like CNN decoder. This alternative is generally weaker on average.

\subsubsection{Optimal binarisation threshold.}
\label{apar:thr}

\begin{figure}
    \centering
    \includegraphics[width=1\linewidth]{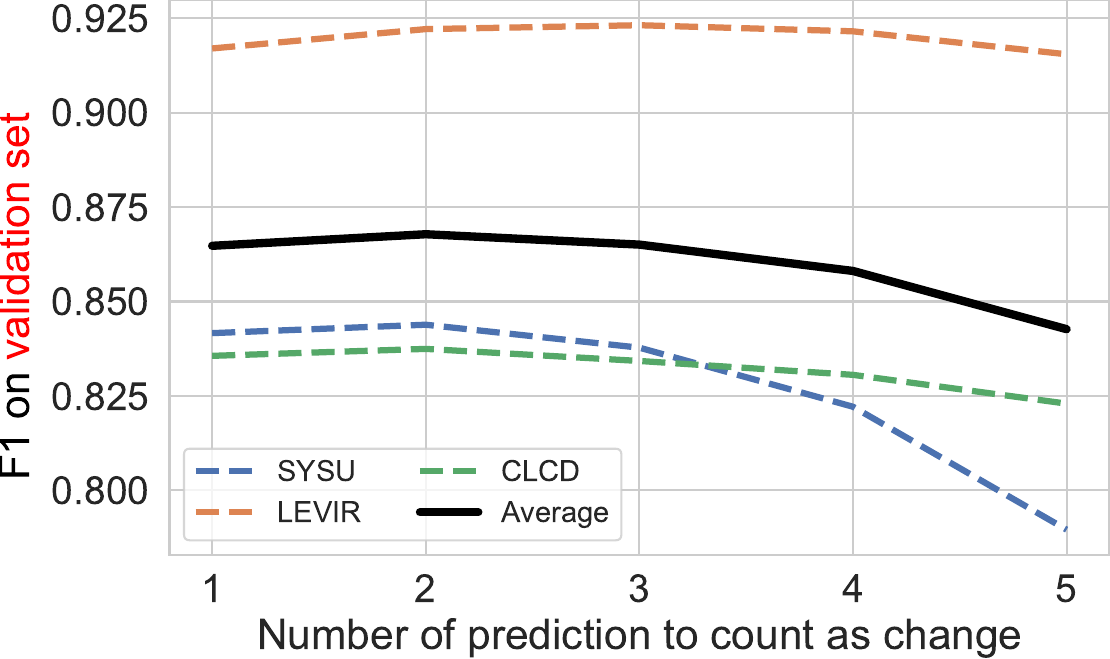}
    \caption{Different thresholds used for binarisation evalauted on \textbf{validataion set}. OSCD does not contain a validation set, so we skip it. The best performance is achieved by binarising all regions where at least two ensemble predictions indicate a change.}
    \label{afig:th_plot}
\end{figure}

\begin{figure}
    \centering
    \includegraphics[width=1\linewidth]{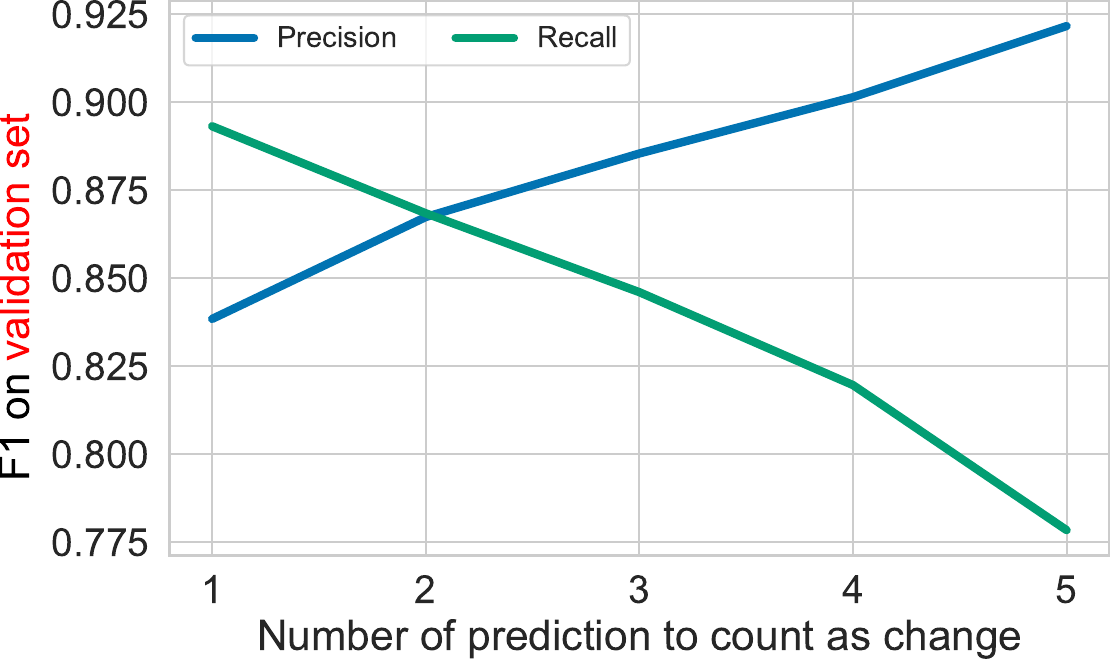}
    \caption{Different thresholds used for binarisation evaluated on \textbf{validation set} (3 datasets average) in terms of precision and recall. OSCD does not contain a validation set, so we skip it. The selection of the binarisation threshold offers a trade-off between recall and precision. The optimal precision-recall trade-off occurs at the 2-predictions-equal-change rule.}
    \label{afig:th_pre_re}
\end{figure}

To find the optimal threshold for binarising our predicted ensemble of change masks, we evaluate different thresholds on \textbf{the validation} set. Since OSCD and SECOND do not contain a validation set, we skip it. \textit{The optimal binarisation regime in an ensemble of five predictions is to consider a region changed if at least two predictions mark it as such}, as shown in \Cref{afig:th_plot}. \Cref{afig:th_pre_re} illustrates that this point represents the best precision-recall trade-off, but the model offers the option to either prefer recall or precision by varying this threshold. For fair evaluation, we use the optimal technique in terms of F1 (2-predictions-equal-change) from the validation set to evaluate ChangeFlow on the test set in the main paper.

\subsection{Additional SCD ablations}
\label{asub:scd_abl}

Here, we present different ablations for various choices in the SCD setting. First, all choices are linked to the model's conditioning and DiT input, then all ablations that are linked to RGB colourisation.

\begin{table}[!ht]
    \centering
    \caption{Ablation of conditioning and construction of input to DiT.}
    \label{tab:scd_cond}
    \begin{tabular}{lccc}
    \toprule
    Method & mIoU & SeK & $F_{scd}$\\
    \midrule
   \rowcolor{gray!20} Ours & 73.9 & 25.3 & 65.9\\
    No $\Delta F$ & 73.9 & 25.0 & 65.3\\
    No bin. $M$ & 73.4 & 24.3 & 64.3\\
    \midrule
    No abs on $\Delta F$ & 73.9 & 25.0 & 65.2\\
    Complex cond. & 73.9 & 25.1 & 65.4\\
    \midrule
    Noise only init. & 73.8 & 23.5 & 63.0\\
    \bottomrule
    \end{tabular}
\end{table}
\subsubsection{Conditioning ablations}

In \Cref{tab:scd_cond} we ablate various options of constructing a conditioning signal for the SCD model, as well as the initialisation of the starting latent for DiT. We first show that both $\Delta F$ and the binary mask $M$ are necessary to achieve the best results (\textit{No $\Delta F$} and \textit{No bin. $M$}) ablations). Of the two, the binary mask offers a better, more compact signal (single channels vs 1024 channels in the case of $\Delta F$), but it does require a separate binary model pass. 

Similarly to BCD, we also evaluate the option without an absolute value on $\Delta F$ (\textit{No abs on $\Delta F$ }) and a complex version that applies a learnable $1x1$ convolutional layer to $\Delta F$ (\textit{Complex cond.}). Both options perform worse than our version. Indicating that, for SCD, order-invariant conditioning (with absolute value) and a simpler model (no learnable layers) offer the best performance.

Finally, we show that replacing the image encodings of VAE perturbed by Gaussian noise ($\big[\,\mathcal{V}(I_1),\; \mathcal{V}(I_2)\,\big] + \epsilon$) with pure Gaussian noise (\textit{Noise only init.}) achieves substantially worse performance. This insight is useful beyond our case, as it can be applied to other dense tasks that operate on 1:1 mappings (e.g., image-to-semantic-mask). Unfortunately, we did not find any benefits of this in the binary case, presumably because it is a 2:1 mapping, as we need to either pick one or combine two image latents, and trivial options do not work there.

\subsubsection{Colour palette ablations}

Our model colourises semantic maps using greedy max-spaced RGB values, which scales to an arbitrary number of classes. \Cref{tab:scd_clr} evaluates two alternatives. The first (\textit{direct}) simply reuses the default RGB colours of the SECOND visualisation palette. Surprisingly, it matches our max-spaced strategy on mIoU (which is dominated by the change vs.\ no-change decision) and is only slightly behind on $F_{scd}$. The second (\textit{corner}) is a binary corner coding: each class is assigned a vertex of the $\{-1,1\}^3$ cube, i.e., every RGB channel takes the minimum or maximum of the normalised VAE input range. Avoiding intermediate values guarantees a pairwise Euclidean distance of at least $2$ between any two class codes. This again performs on par with \textit{direct} and slightly below our strategy. Overall, performance is robust to the colourisation scheme, provided that the class codes are well-separated.

\begin{table}[!ht]
    \centering
    \caption{Ablation of colour palette used to encode semantic change masks into an RGB image.}
    \label{tab:scd_clr}
    \begin{tabular}{lccc}
    \toprule
    Method & mIoU & SeK & $F_{scd}$\\
    \midrule
   \rowcolor{gray!20} Greedy max-spaced (Ours) & 73.9 & 25.3 & 65.9\\
    Direct & 73.9 & 25.1 & 65.6\\
    Corner & 73.9 & 25.1 & 65.5\\
    \bottomrule
    \end{tabular}
\end{table}

\section{Additional qualitative results}
\label{a:qual}

This section provides additional qualitative results. First, we present the extended main qualitative results in comparison with a wider selection of related work (\Cref{asub:main_q}), followed by additional SCD qualitative results (\Cref{asub:scd_q}), and failure cases (\Cref{asub:fail_q}). Next, we present both binary and semantic visual VAE mask reconstructions (\Cref{asub:vae_q}) and visualisations of intermediate generation steps (\Cref{asub:gen_q}).

\subsection{Main qualitative results}
\label{asub:main_q}

\begin{figure*}
    \centering
    \includegraphics[width=1\linewidth]{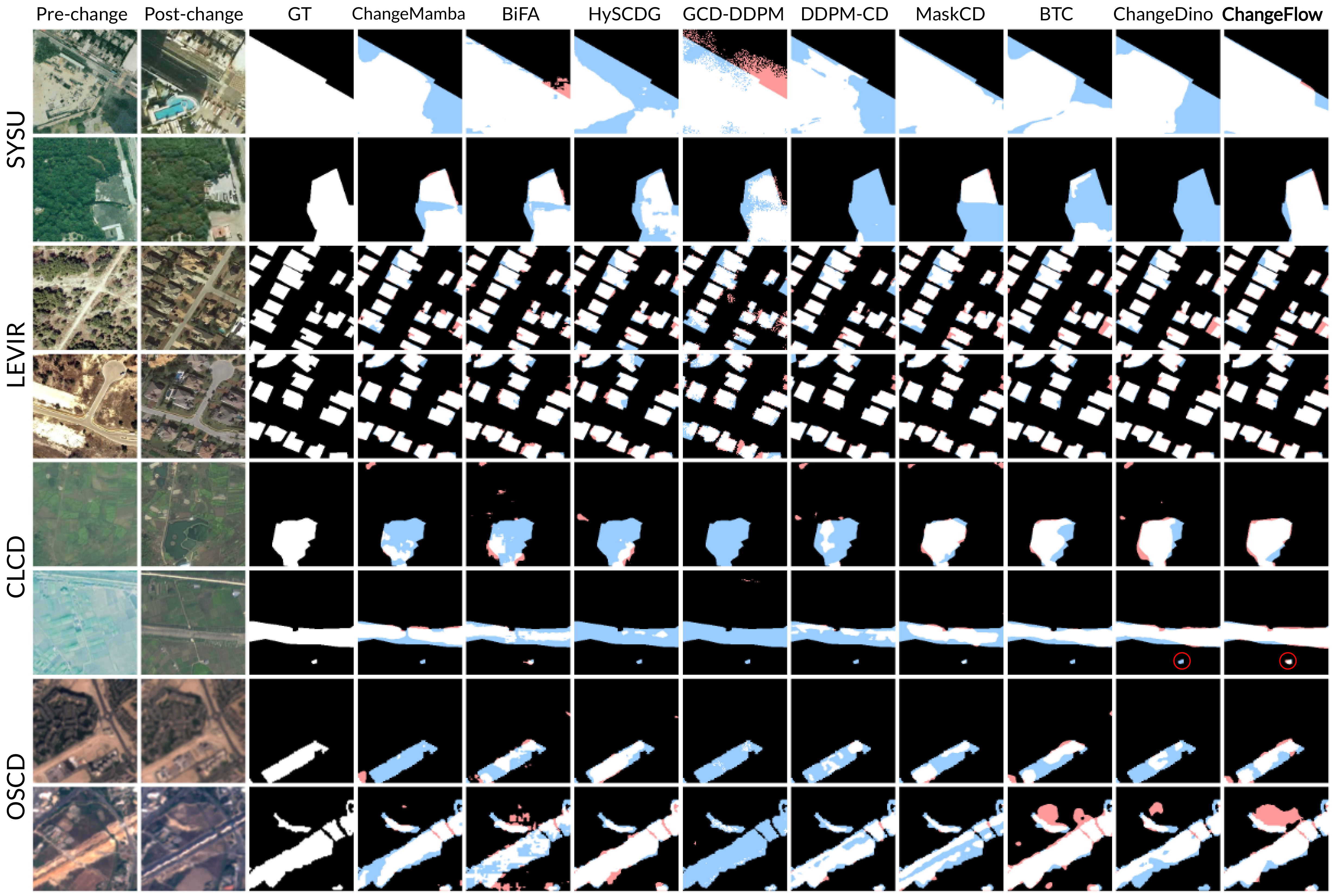}
    \caption{Additional qualitative results.}
    \label{afig:qual}
\end{figure*}

\Cref{afig:qual} presents visual results in comparison to a wider set of related methods. ChangeFlow excels at predicting more coherent change masks and capturing full changed regions (low number of false negatives). No prior method can consistently match this behaviour across multiple datasets, as also reflected in ChangeFlow's superior recall (see \Cref{a:ext_main}).

\subsection{Additional SCD results}
\label{asub:scd_q}

\begin{figure}
    \centering
    \includegraphics[width=1\linewidth]{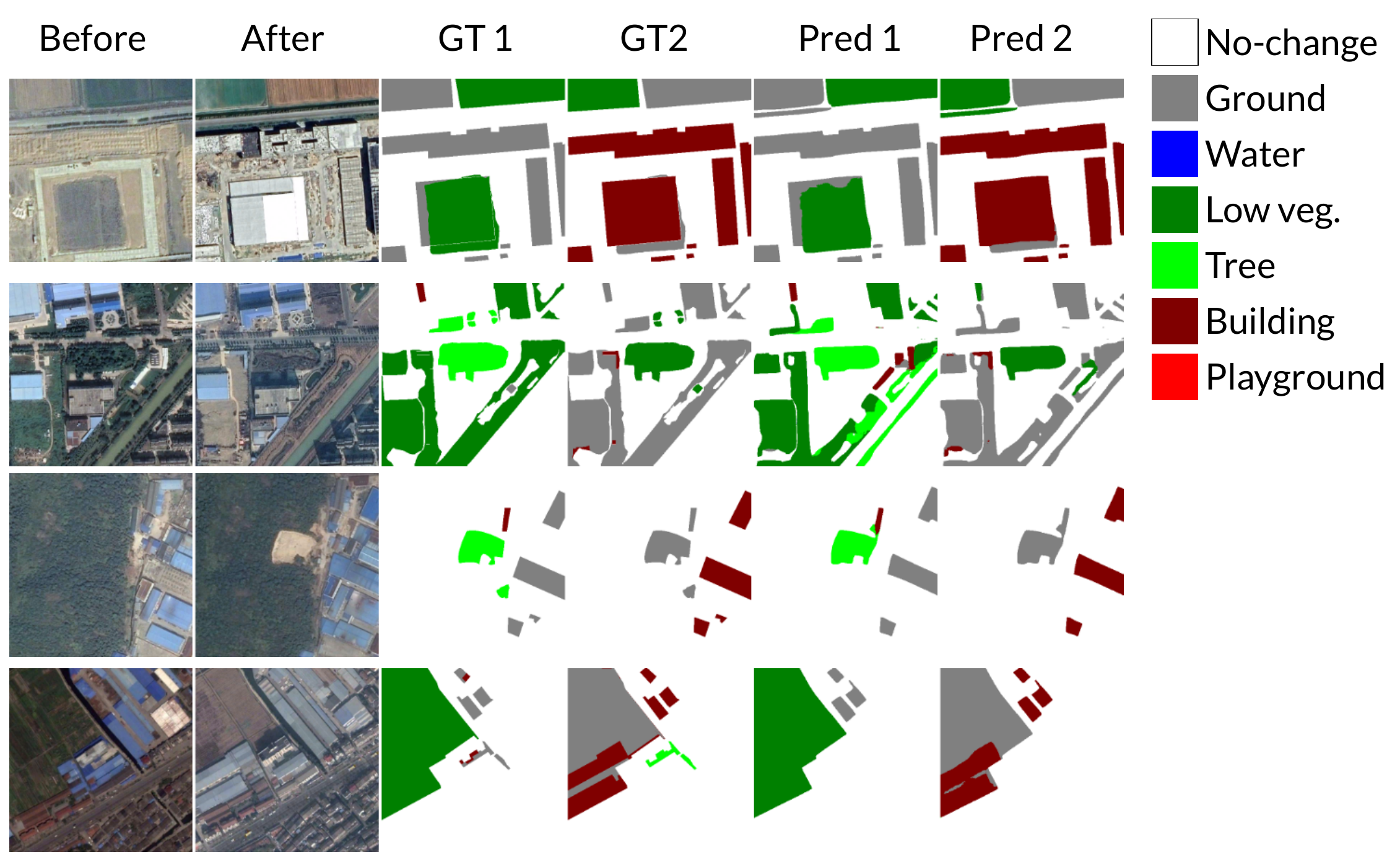}
    \caption{Additional SCD qualitative examples.}
    \label{afig:scd}
\end{figure}

Additional qualitative examples in the SCD setting evaluated on SECOND are shown in \Cref{afig:scd}.

\subsection{Failure cases}
\label{asub:fail_q}

\begin{figure*}
    \centering
    \includegraphics[width=1\linewidth]{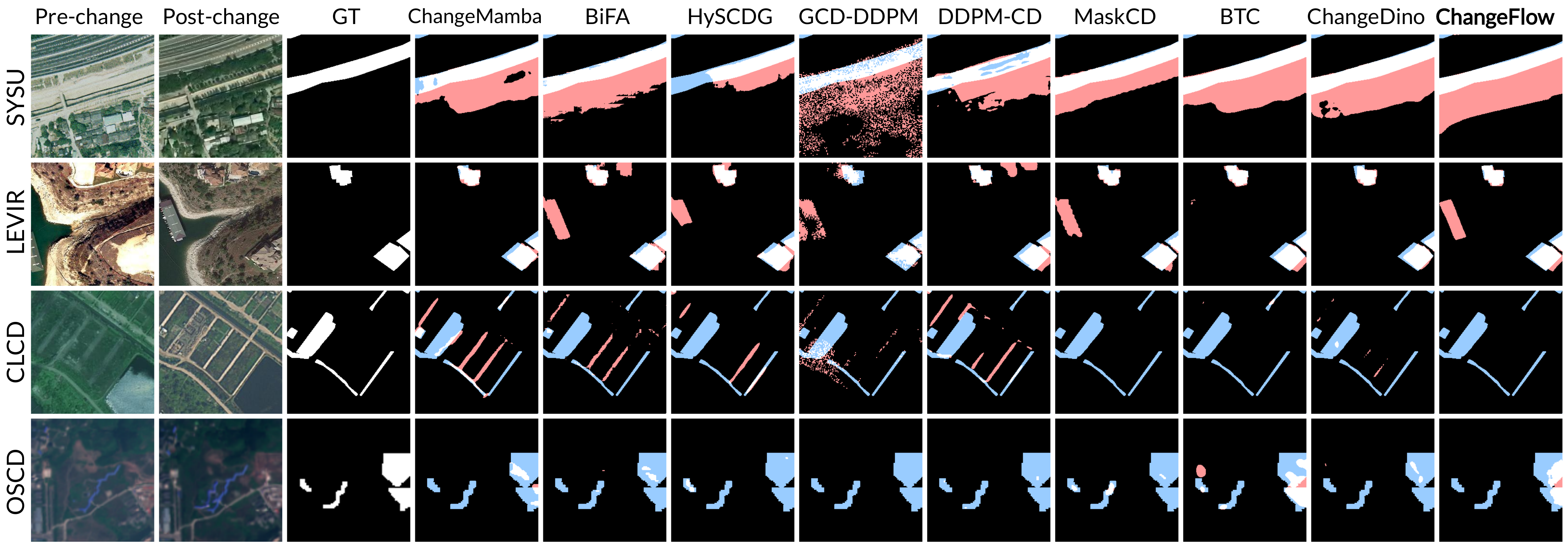}
    \caption{Additional failure qualitative results.}
    \label{afig:fail}
\end{figure*}

\Cref{afig:fail} contains a visualisation of some failure cases. ChangeFlow does miss some changed regions in specific situations, but the visualisations show that most other methods struggle with similar problems. The hardest example is shown in the CLCD row, where no model correctly predicts the majority of the changed region, indicating its high semantic nature and difficulty. In the first two rows (SYSU and LEVIR), we see that some of these changes may be due to mislabelling. Later (see additional confidence visualisations below), we show that the model is quite uncertain about the misclassified LEVIR case.

\subsection{Visualisation of VAE mask reconstruction}
\label{asub:vae_q}

\begin{figure}
    \centering
    \includegraphics[width=1\linewidth]{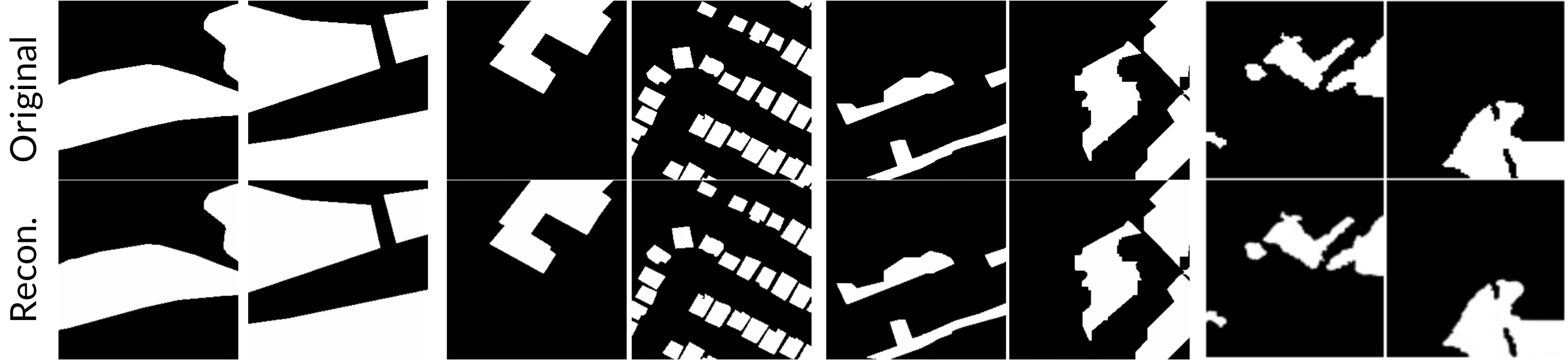}
    \caption{Original (top row) and reconstruction (bottom row) of binary change masks through pretrained SD-XL VAE. The encoding process preserves the details and structure of masks with minimal data loss. Refer to the main paper for quantitative evaluation.}
    \label{afig:vaeRecon}
\end{figure}

Our method uses a pretrained variational autoencoder (VAE) from SD-XL~\cite{podell2024sdxl}. This network was originally trained on RGB images, so it is immediately obvious that we can also encode binary change masks with minimal loss of data. We verified this and presented the results in the main paper, with minimal drop in F1 score, BF1 score, and mean absolute error. In \Cref{afig:vaeRecon}, we also support the quantitative results with visual proof that the VAEs trained for RGB images sufficiently encode binary masks. The implementation details for the analysis are in \Cref{asub:abl_imp}. Even if this whole observation might not be intuitive at first glance, we note that a binary image is a special case of an RGB image, so it is natural that a certain subspace of the VAE latent space encodes such imagery.

\begin{figure}
    \centering
    \includegraphics[width=1\linewidth]{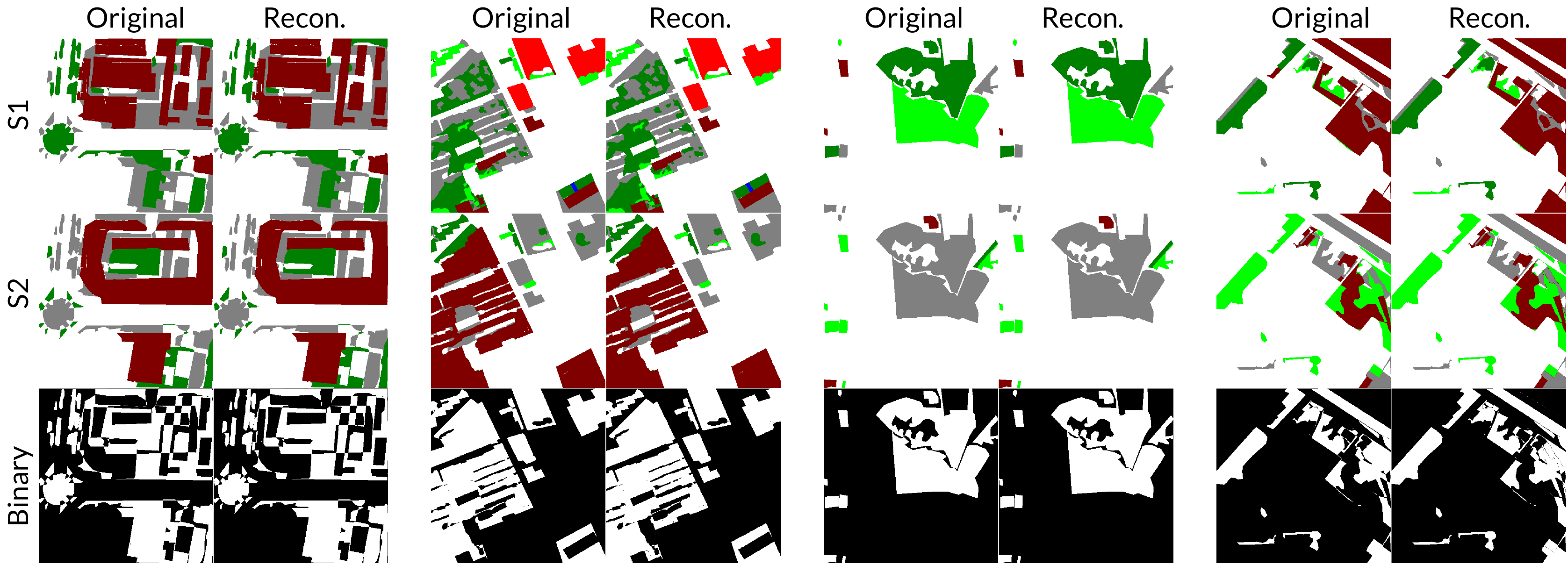}
    \caption{Original (left column) and reconstruction (right column) of SECOND semantic and binary change masks through pretrained SD-XL VAE. The encoding process preserves the details and structure of masks with minimal data loss. Refer to the main paper for quantitative evaluation.}
    \label{afig:scdRecon}
\end{figure}

We also visually verify the minimal data loss of RGB colouring and VAE pass-through (encode, decode) in \Cref{afig:scdRecon}. As verified by the quantitative results in the main paper, there are only a few errors, even visually.

\subsection{Intermediate steps visualisation}
\label{asub:gen_q}

\begin{figure}
    \centering
    \includegraphics[width=1\linewidth]{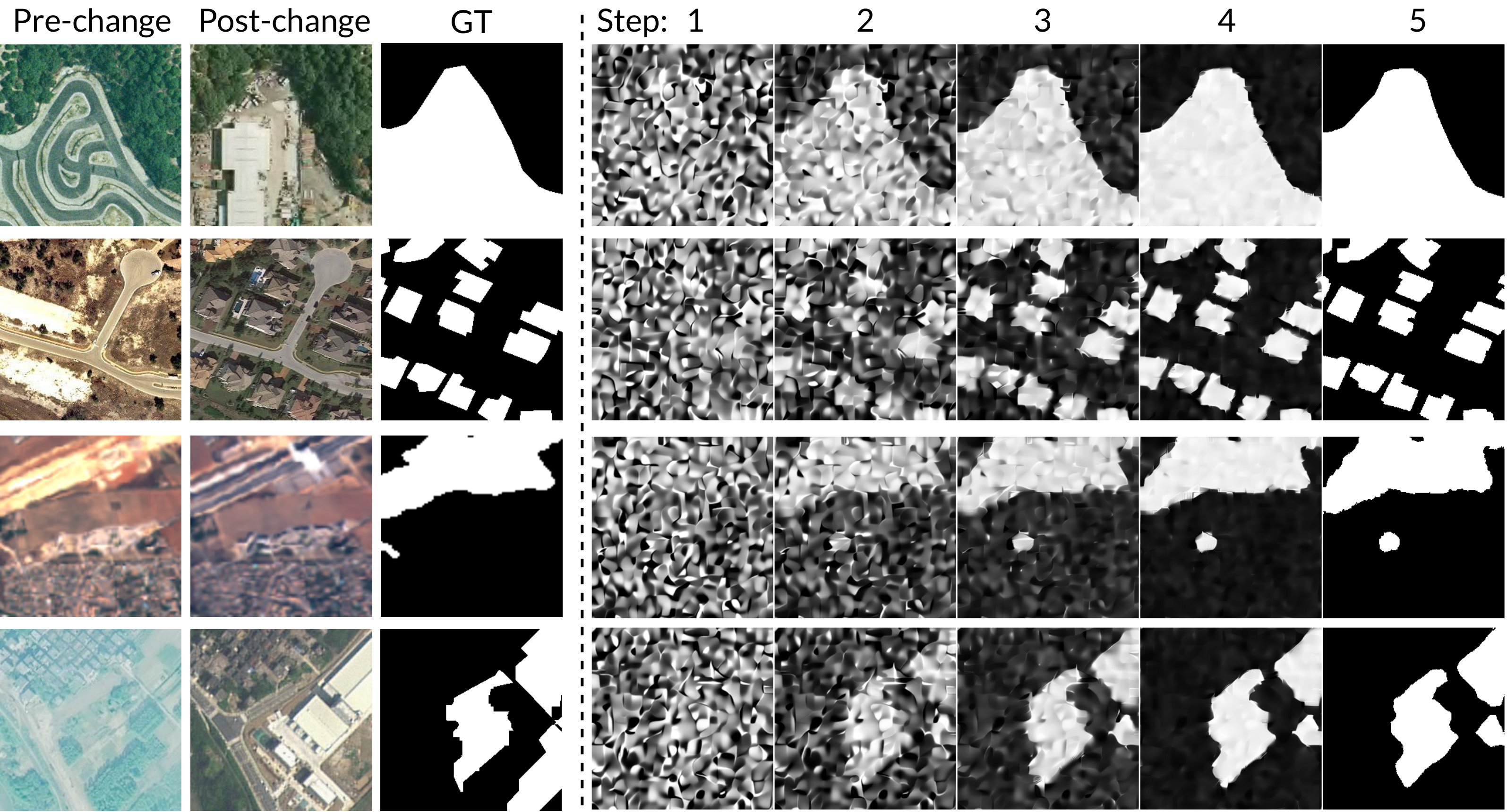}
    \caption{Visualisation of intermediate steps in the latent generative mask prediction. ChangeFlow iteratively predicts from pure noise to a binary mask. Here, we decode the intermediate latent representation into a binary mask at each step. }
    \label{afig:steps}
\end{figure}

\Cref{afig:steps} presents a visualisation of 5 generation steps. As we can see, and as we have quantitatively evaluated in the main paper, the coherent region appears quite early in the process. The final steps then predominantly focus on border refinement (we recommend zooming in to see this clearly).

\section{Computational efficiency}
\label{a:comp}
Details of our computational-efficiency evaluation protocol are provided in \Cref{asub:per_prot}. We additionally report complementary metrics for our method and selected state-of-the-art baselines, including GFLOPs and inference time in \Cref{asub:addit_comp_res}.

\subsection{Protocol}
\label{asub:per_prot}

We report three efficiency metrics: parameter count, inference time (also expressed as frames per second, FPS), and GFLOPs. GFLOPs are measured with the official PyTorch profiler.

The protocol closely follows the one from~\cite{rolih2025btc}. Inference time is measured using a pair of $256 \times 256$ RGB inputs. All models, except ours, are evaluated in \texttt{float16} where supported. Our model does not support float16 for all modules; therefore, we use torch compile when measuring inference time. To robustly measure the metrics, we perform 1000 warm-up forward passes followed by 1000 timed forward passes; this procedure is repeated five times, and we report the average runtime per forward pass. All measurements are conducted on an NVIDIA A100-SXM4 40GB GPU (and AMD Epyc 7H12 CPU) on a Slurm cluster. We use the same protocol for all methods and efficiency results.

\subsection{Extended computational results}
\label{asub:addit_comp_res}

\Cref{atab:perf} shows extended computational efficiency results including frames per second (FPS) derived from inference time, parameter count, floating operations per second (FLOPS), and change detection metrics. ChangeFlow outperforms ChangeDINO by 1.3 points in change detection but offers comparable throughput and inference time. Some other discriminative methods, such as BTC and HySCDG, do achieve faster inference, but they do not perform nearly as well as ChangeFlow in change detection. 

Due to multiple inferences, ChangeFlow has a higher FLOPS value, in line with related diffusion-based models (DDPM-CD and SatDiFuser). \textit{This is not an issue on modern hardware with massive parallelisation capabilities}, as reflected by FPS metric. We argue that FPS is a much better measure of the model's actual inference speed, as reflected in the comparison between our 1step5rep and 5step1rep setups: the FLOPS value is higher (due to parallelisation) for the multiple-repetition setup, but inference is actually faster, as these repetitions can be easily parallelised on modern GPUs.

The previous generative method, GCD-DDPM, is almost 3 orders of magnitude slower than our method. This stems from its complex conditioning scheme, which uses the auxiliary CD architecture's output for guidance. It also operates in pixel space and does 1000 generation steps, as opposed to 5 latent steps in ChangeFlow. 

Another observation is that, even though ChangeFlow is a true generative method, it operates faster than DDPM-CD and SatDiFuser, which use a generative (diffusion) network solely as a feature extractor, while achieving substantially higher change detection accuracy.

\begin{table}[!ht]
    \centering
        \caption{Computational efficiency results for each model. We report FPS (derived from inference time), inference time, parameter count, GFLOPs, and average Precision, Recall, and F1 across 4 datasets. All results were obtained using an Nvidia A100-SXM4 40GB GPU using the above-described protocol.}
        \label{atab:perf}
        \setlength{\tabcolsep}{3pt}
     \resizebox{1\linewidth}{!}{
    \begin{tabular}{lcccccccccccccccccc}
    \toprule
    	\multirow{2}{*}{~} & FPS & Inference Time & Param. & FLOPS& \multicolumn{3}{c}{\textit{Avg}} \\
~ & [img/s] & [ms] & [M] & [$10^9$]& Pr. & Re. & F1\\
 \hline
FC-Siam-Diff~\cite{daudt2018fcn} & 170.1{\scriptsize$\pm0.0$} & 5.9{\scriptsize$\pm0.0$} & 1.4 & 4.6 & 62.1 & 66.2 & 61.5\\
ChFormer~\cite{bandara2022changeFormer} & 36.2{\scriptsize$\pm0.0$} & 27.6{\scriptsize$\pm0.0$} & 41.0 & 234.6 & 74.0 & 65.3 & 69.1\\
SwinSUNet~\cite{zhang2022swinsunet} & 33.1{\scriptsize$\pm0.1$} & 30.2{\scriptsize$\pm0.1$} & 43.6 & 32.6 & 79.3 & 69.4 & 73.6\\
GFM~\cite{mendieta2023gfm} & 44.9{\scriptsize$\pm0.1$} & 22.3{\scriptsize$\pm0.1$} & 120.5 & 109.2 & 79.6 & 72.2 & 75.7\\
GCD-DDPM~\cite{wen2024gcd-ddpm} & 0.02{\scriptsize$\pm0.0$} & 43563.6{\scriptsize$\pm0.0$} & 131.9 & 531997.7 & 56.1 & 54.4 & 49.8\\
BiFA~\cite{zhang2024bifa} & 32.2{\scriptsize$\pm0.0$} & 31.0{\scriptsize$\pm0.0$} & 9.9 & 4.3 & 79.8 & 66.4 & 71.3\\
MaskCD~\cite{yu2024maskcd} & 6.5{\scriptsize$\pm0.0$} & 153.5{\scriptsize$\pm1.0$} & 107.4 & 143.2 & 80.0 & 66.8 & 71.4\\
ChMamba~\cite{chen2024changeMamba} & 14.4{\scriptsize$\pm0.0$} & 69.6{\scriptsize$\pm0.2$} & 92.4 & 96.2 & 83.2 & 69.1 & 74.9\\
MTP~\cite{wang2024mtp} & 31.2{\scriptsize$\pm0.0$} & 32.1{\scriptsize$\pm0.0$} & 107.8 & 196.9 & 77.6 & 77.0 & 76.5\\
HySCDG~\cite{benidir2025hyscdg} & 41.0{\scriptsize$\pm0.1$} & 24.4{\scriptsize$\pm0.0$} & 65.1 & 64.8 & 77.8 & 67.2 & 71.9\\
DDPM-CD~\cite{bandara2025ddpmcd} & 4.6{\scriptsize$\pm0.0$} & 217.6{\scriptsize$\pm0.4$} & 437.5 & 8871.2 & 80.3 & 63.8 & 70.0\\
SatDiFuser~\cite{jia2025satdifuser} & 1.8{\scriptsize$\pm0.0$} & 542.2{\scriptsize$\pm0.9$} & 1413.6 & 6142.9 & 84.7 & 70.8 & 76.6\\
BTC~\cite{rolih2025btc} & 32.4{\scriptsize$\pm0.0$} & 30.8{\scriptsize$\pm0.0$} & 120.1 & 221.4 & 83.3 & 72.3 & 77.3\\
ChangeDINO~\cite{cheng2025changedino} & 8.9{\scriptsize$\pm0.5$} & 112.3{\scriptsize$\pm6.2$} & 311.1 & 1269.1 & 85.2 & 74.4 & 79.1\\
\textbf{ChangeFlow}\tiny{(5step, 5rep)} & 11.8{\scriptsize$\pm0.0$} & 84.6{\scriptsize$\pm0.0$} & 403.3 & 4673.9 & 81.7 & 79.2 & 80.4\\
\midrule
ChangeFlow \tiny{(1step, 5rep)} & 18.7{\scriptsize$\pm0.0$} & 53.5{\scriptsize$\pm0.0$} & 403.3 & 3543.1 & 82.5& 78.4& 80.3\\
ChangeFlow \tiny{(5step, 1rep)} & 33.8{\scriptsize$\pm0.1$} & 29.6{\scriptsize$\pm0.5$} & 403.3 & 1188.4 & 82.9& 77.1& 79.8\\
    \bottomrule
    \end{tabular}
    }
\end{table}

\section{Extended implementation details}
\label{a:ext_impl}

This section contains detailed implementation details for our model in \Cref{asub:our_impl}, implementation details for our ablations and analysis (coherence) in \Cref{asub:abl_imp}, and finally also details regarding the related methods in \Cref{asub:rel_impl}.

All experiments, including all our experiments and related method execution, were conducted on an NVIDIA A100-SXM4 40GB GPU (and AMD Epyc 7H12 CPU) on a Slurm cluster.

\subsection{Our model - ChangeFlow}
\label{asub:our_impl}

This subsection contains additional implementation details not included in the main paper for the modules used in ChangeFlow: the SD-XL VAE, the diffusion transformer (DiT), the DINOv3 image encoder, feature difference and normalisation, ensembling details, extension to SCD details and other training-related details.

\subsubsection{SD-XL VAE.}
We use the VAE from SD-XL (Stable Diffusion XL)~\cite{podell2024sdxl} for image generation. It is kept frozen in the base model, so no gradient flows through the encoder or the decoder. We selected it for its compact 4-channel latent space ($d=4$). We also ablated this choice as useful in ablation studies of the main paper.
Specifically, we use the HuggingFace \textcolor{weights}{stabilityai/sdxl-vae} version and keep all details unchanged. We utilise the scaling factor and, as standard~\cite{podell2024sdxl}, multiply the latent by 0.13025 and then, before decoding, divide by it. As already explained in the main paper, to convert a binary mask to RGB, we simply repeat the value along the channel dimension. When decoding, we simply average the 3 RGB channels to retrieve a single-channelled binary mask. Since the VAE expects images to be normalised to the range $[-1, 1]$, we rescale all masks to this range before encoding, then back to $[0, 1]$ after decoding. For SCD semantic change mask coloring refer to \Cref{asub:scd_impl}.

\subsubsection{DiT model.}
As the model that predicted the velocity field, we opt for the recent diffusion transformer (DiT). The architecture itself is based on \textcolor{weights}{LLaMA-2 DiT}, with the implementation adopted from the \textcolor{weights}{minRF} GitHub repo. We set the channel dimension to 256, use 10 layers with 8 heads each, and a patch size of 1.  We do not use class embeddings or classifier-free guidance. The input channel dimension is set to the sum of the image encoder dimension $c$ and the VAE latent dimension $d$, specifically 1024 + 4, for a total of 1028, since the model receives a concatenation of feature difference and noise in the shape of a mask VAE latent. Output channel dimension is set to VAE latent dimension (4, which matches the latent of the expected output mask latent). The model also takes the time-step value in the range $[0, 1]$ as an input, which is then embedded using the TimeEmbedder (see the repo mentioned above for details). Other DiT hyperparameters remain the same as in the repo mentioned above. This module has an initial learning rate of $1\cdot 10^{-4}$.

\subsubsection{Image encoder.}

We use DINOv3 ViT-L as a shared weight image encoder, specifically the version from HuggingFace \textcolor{weights}{facebook/dinov3-vitl16-pretrain-lvd\-1689m}. It has a hidden dimension of 1024, a patch size of 16, and 24 layers. We do not modify any default hyperparameters and use the provided image normalisation parameters. The model is finetuned during training following~\cite{rolih2025btc}, and we set the learning rate of this module to $5\cdot10^{-5}$. We extract the features from the last (24th) layer. We discard register and class tokens and reshape features from $\mathbb{R}^{l \times c}$ to $\mathbb{R}^{h' \times w' \times c}$, where $h' = w' = \sqrt{l}$ and $c=1024$.

\subsubsection{Feature difference and normalisation.}

To obtain a conditioning vector, features extracted with the above-described image encoder are normalised before subtraction (differencing). We opt for LayerNorm~\cite{ba2016layer}, a standard choice and the best performer according to ablations in \Cref{asub:norm_ab}. This is applied across channel (embedding) dimension $c$, in the feature map $\mathbb{R}^{h' \times w' \times c}$. LayerNorm hyperparameters remain default as in PyTorch, and the trainable scale parameters have the same learning rate as DiT: $1\cdot 10^{-4}$.

Feature difference is computed per element, meaning that given two feature maps, both of shape $\mathbb{R}^{h' \times w' \times c}$, we subtract the values at the same indices of $h',w', c$. In our base model, we then apply the absolute value to this difference. This absolute difference represents our conditioning vector.

Finally, the conditioning vector is resized to match the VAE spatial dimensions (in our case, we upscale $ h'$ and $ w'$ by a factor of 2) using simple bicubic interpolation (PyTorch implementation). This choice of resizing method is ablated in \Cref{asub:cond_res}. 

\subsubsection{Ensembling details.}

To obtain an ensemble of predictions, we repeat the inference $N$ times, where $N=5$ in our base model, the same way as explained for a single inference. More specifically, this means that we sample $N$ noise vectors $\mathbb{R}^{h \times w \times d} \sim \mathcal{N}(0,1)$ that have the same shape as the expected mask VAE latent. These then undergo the standard 5-step inference via ODE integration, resulting in $N$ final change-mask latents: $\{\hat{x}_i | i\in0..N\}$. These are then individually decoded via the VAE decoder, and the RGB channels are averaged to obtain single-channel binary masks, resulting in a final ensemble of binary change masks $\{\hat{M}_i | i\in0..N\}$. We then stack predictions in a new dimension to obtain $ \hat{M}_{ens} \in \mathbb{R}^{N \times h \times w}$ and aggregate via averaging across the new dimension to obtain a final prediction $\hat{M} \in\mathbb{R}^{h \times w}$.

During inference, this process can be easily parallelised since the \textit{repetitions are independent}. By stacking the $N$ different initial noise vectors $\{x_o^i|i \in 0..N \}$ in a new dimension to get $ x_o^{batched} \in \mathbb{R}^{N \times h \times w \times d}$, the ODE integration is performed in batched manner, resulting in a batched final latent $\hat{x}^{batched} \in \mathbb{R}^{N \times h \times w \times d}$, which is then decoded and merged as explained above. This means that increasing repetitions increases inference time with respect to the parallelisation capabilities of modern hardware, in theory enabling a smaller overhead with better parallelisation.

\subsubsection{Semantic change detection details}
\label{asub:scd_impl}

\begin{figure*}
    \centering
    \includegraphics[width=1\linewidth]{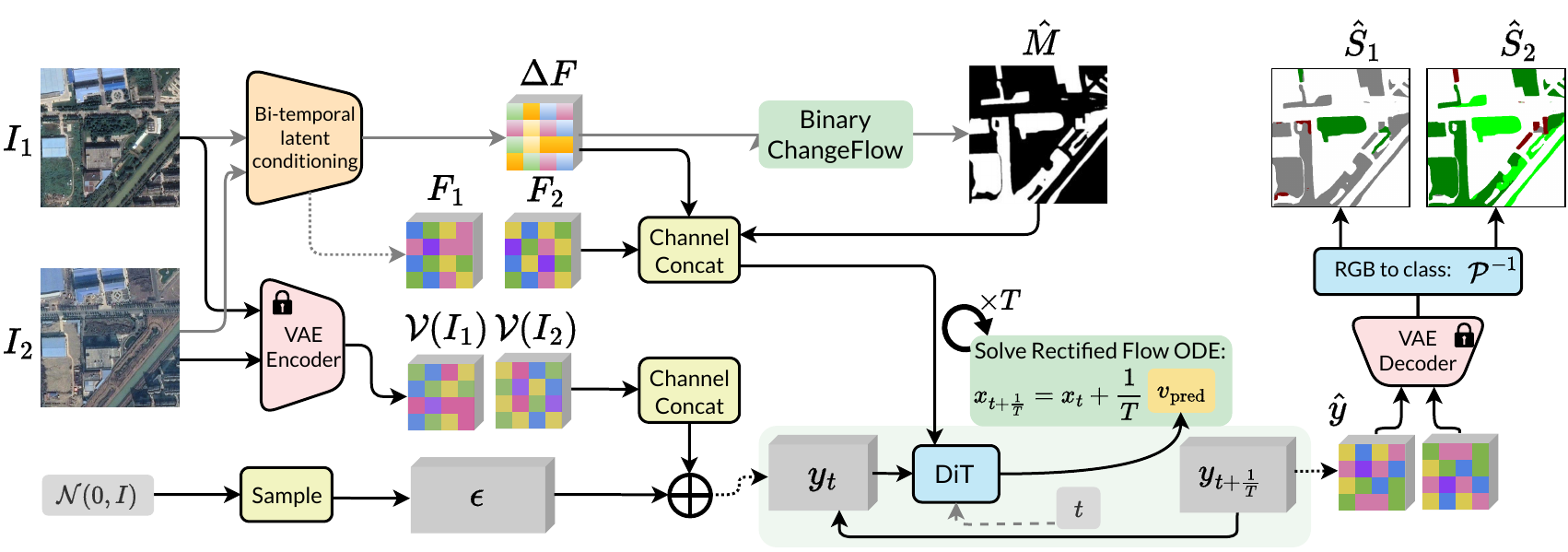}
    \caption{Semantic change detection flow diagram. Unlike BCD, the initial latents consist of VAE encoded images perturbed by Gaussian noise. The model takes this as an input, as well as conditioning, which is a channel-concatenated binary mask $M$, raw encoder features $F_1, F2$, and their absolute difference $\Delta F$. The model then predicts the final latents by integrating the velocity field, as in BCD. Final latents are first decoded using a frozen VAE, and the RGB values are then decoded to class indices (to obtain semantic change maps $S$) via a nearest-neighbour lookup in the colour palette ($\mathcal{P}^{-1}$).}
    \label{afig:scd_arch}
\end{figure*}

\noindent\textbf{(i) Semantic mask encoding (RGB colouring).}
As explained in the main paper, we map the semantic classes of the land-cover maps $S_1, S_2$ to RGB colours with a fixed palette $\mathcal{P}$, so both maps can be processed by the same frozen SD-XL VAE. The palette is constructed once with a greedy max-distance algorithm: the no-change class is fixed to white $(255, 255, 255)$, and the remaining $K{=}6$ SECOND classes are assigned iteratively, each time selecting the candidate RGB colour that maximises the minimum Euclidean distance to all previously selected colours. Candidates close to the grey axis (i.e., with near-equal R, G, and B values) are excluded, since low-saturation colours are the most susceptible to confusion after the VAE round-trip. The resulting colours are maximally separable in RGB space, which makes the inverse mapping trivial: a decoded RGB prediction is converted back to class indices by nearest-neighbour assignment of each pixel to the closest palette entry ($\mathcal{P}^{-1}$ in the main paper). As reported in the main paper, encoding--decoding semantic masks this way is nearly lossless ($F_{scd}$ of 99.45 and mIoU of 99.53), and performance is robust to the exact palette (as shown in ablations in \Cref{asub:scd_abl}). Both colourised maps $C_1 = \mathcal{P}(S_1)$ and $C_2 = \mathcal{P}(S_2)$ are encoded independently by the VAE and channel-concatenated into a single target latent $y_1$, so both timestamps are generated jointly.

\noindent\textbf{(ii) SCD Conditioning.}
The SCD model is conditioned by channel-concatenating the following signals to the interpolated latent $y_t$: the binary change mask $M$, the bi-temporal features $F_1$ and $F_2$ (from 21st layer), and normalised absolute feature difference $\Delta F$ (computed exactly as in the binary case from last layer). The raw features $F_1, F_2$ from the encoder provide semantic guidance for the two land-cover maps, while $M$ provides binary change guidance. All conditioning signals are bicubically resized to the latent resolution, and the DiT input channel dimension is adjusted accordingly (same as in the binary case). During training, $M$ is the ground-truth binary mask; during inference, it is replaced by the prediction $\hat M$ of a separately trained binary ChangeFlow model. The same $\hat M$ is reused a second time at the output: the decoded semantic maps are gated with $\hat M$, setting all pixels outside the predicted change region to the no-change class. These choices are ablated in the supplementary material above.

\noindent\textbf{(iii) Other SCD details.}
Unlike the binary case, the rectified-flow source is not pure noise: the source $y_0$ is the channel-concatenation of the two \emph{image} latents, perturbed with Gaussian noise $\epsilon \sim \mathcal{N}(0, I)$, which preserves the image prior while retaining the stochasticity required for sampling-based ensembling. We also tried some of these variations in a binary setting, where this is not as simple since we need to combine two image latents into a single $x_0$. We found that it offers no improvements in the binary case. The semantic case is ablated in the supplementary material above.

Timestep sampling during training follows a distribution that is more concentrated towards the initial time step, as proposed by~\cite{yang2026genmask}. Inference likewise mirrors the binary case: Euler integration over $T=5$ equally spaced steps, starting from $y_0$, with an ensemble aggregated by per-pixel majority vote over the predicted class indices (after nearest-neighbour decoding), rather than by averaging as in the binary case. Input images are $512 \times 512$ following the standard SECOND protocol. The SCD model is trained for 150 epochs with a batch size of 8 (due to the larger image size). All remaining hyperparameters (optimiser, learning rates, scheduler, and augmentations) are identical to the binary configuration described above.

\subsubsection{Other details.}

As already explained in the main paper and above, we obtain a single channel prediction from the ensemble of $N$ predictions $\mathbb{R}^{N\times h\times w}$ by averaging across the ensemble dimension $N$ in a binary setting to get $\mathbb{R}^{h \times w}$. The values in this prediction are continuous but represent 5 different hypotheses. To achieve the effect of predictions indicating a change in continuous space, we set the threshold to 0.3. This is equivalent to discretising into 5 values (with rounding) and then thresholding at $\geq 2$. This was established as the optimal threshold on the validation set with results presented in \Cref{apar:thr}. In SCD no thresholding is required as we use majority voting.

As explained in the main paper, we sample time steps during training in a logit-normal fashion in BCD (see \Cref{asub:time_abl} for details and ablations). In SCD, we use GenMask~\cite{yang2026genmask} inspired sampling. During inference, time steps are equally spaced on the interval $[0, 1]$. In our case, we use 5 timesteps ($T=5$). The value was simply chosen as the one where performance, according to ablations, is good, but the same ablation also shows that increasing $T$ does not yield consistent gains. The number of repetitions in the ensemble (i.e., 5) was selected as it represents a good speed-performance trade-off. While we could've selected a higher value to achieve even better CD performance, we believe our choice is fair, given that its inference speed is similar to the previous best method.

As already explained in the main paper, we use rotation and flipping augmentations, each applied with a probability of $30\%$. All input images in binary setting are of size $256\times256$, which means that for OSCD, we rescale the images from crops of $96\times96$, following other works~\cite{wang2024mtp, rolih2025btc}. For the SECOND we follow the original dataset where images are $512\times512$ pixels. Data normalisation is specified above for the image encoder and VAE. Dataset details are in \Cref{a:data}.

A cosine scheduler without restarts is used in all cases, with the PyTorch default implementation. The Muon optimiser comes from the Timm library. We picked this option with the recent success of LLM applications, but the change in results compared to AdamW was minimal in preliminary studies.

Metric implementations come from TorchMetrics and augmentations from Albumentations.

\subsection{Ablation and analyses implementation details}
\label{asub:abl_imp}

\subsubsection{Encoder ablation details.}

All parameters stay the same as for the base model, except for the following, which are specific to encoder selection.
For DINOv2, we use \textcolor{weights}{facebook/dinov2-large}; all hyperparameters stay the same. For the DINOv3 satellite, we use \textcolor{weights}{facebook/dinov3-vitl16-pretrain-sat493m}, all hyperparameters stay the same. For RADIO 2.5, we use \textcolor{weights}{nvidia/RADIO-L}; all hyperparameters stay the same, except the learning rate, which is divided by 10, and normalisation is set to author-provided. For RADIO 4, we use \textcolor{weights}{nvidia/C-RADIOv4-SO400M} (ShapeOptimised version since there is no ViT-L), and the hyperparameters are the same as in RADIO 2.5.

\subsubsection{Conditioning ablation details.}

The process of base-feature normalisation and differencing is explained in the implementation details above for ChangeFlow. For other ablated options, we list the details here.

Feature difference is computed per element, meaning that given two feature maps, both of shape $\mathbb{R}^{h' \times w' \times c}$, we subtract the values at the same indices of $h',w', c$. When we compute a signed difference, we subtract the feature map of the second image (the one at a later time step) from that of the first. In the case of concatenation conditioning vector, we concatenate the features in the channel dimension to obtain $\mathbb{R}^{h'\times w'\times 2c}$ (and accordingly adjust the DiT input channel dimension).

In the case of L2 normalisation, we compute the L2 vector norm across the channel dimension of feature map $\mathbb{R}^{h' \times w' \times c}$ (resulting in $\mathbb{R}^{h'\times  w'}$ norm vector), then divide all corresponding channel values by this norm. Unlike LayerNorm, this option does not contain the learnable scale parameters.

\subsubsection{Discriminative ChangeFlow ablation details}

The setup used for the discriminative model is architecturally exactly the same as our generative model. The only difference is that we do not use the generative loss, but replace it with the Dice loss~\cite{Milletari2016diceVnet} on the final VAE decoded mask. The VAE is frozen, but it propagates gradients back to the DiT and the encoder. The model does a 1-step prediction in this case.

\subsubsection{VAE ablation details.}

We use the following VAEs from Huggingface and leave all hyperparameters the same as the original: \textcolor{weights}{stabilityai/sdxl-vae}, \textcolor{weights}{stabilityai/\-stable-diffusion-3.5-medium}, \textcolor{weights}{black-forest-labs/FLUX.1-dev}, and \textcolor{weights}{Tongyi-MAI/Z-Image-Turbo}. The encoder is always frozen, while the decoder is frozen except in ablations where indicated. Since the input of DiT is defined as the sum of the VAE latent dimension $d$ and the image encoder latent dimension $c$, the VAE part of the dimension is accordingly changed to the latent dimension of VAE: $d=4$ for SD-XL and $d=16$ for all others. 

In experiments where we also finetune the SD-XL VAE decoder, we use a standard binary dice loss (same as in \cite{rolih2025btc}) on the change mask, computed with single-step single-repeat inference and binarised from RGB to a single channel. The gradient passes through both the VAE decoder and the DiT, enabling us to avoid the standard rectified flow MSE loss in the "Pixel loss only" experiments. All other parts of DiT and the image encoder keep the same configuration in these experiments. We set the VAE decoder learning rate to the same as DiT's: $1\cdot 10^{-4}$. In the case of the CNN decoder, it is a UNet-like model with a single final CNN block with a channel dimension of 256. Its weights are randomly initialised. Even with the CNN decoder, we keep the VAE encoder for target mask encoding. The learning rate, gradient propagation, and loss are the same as in the finetuned VAE case explained above.

\subsubsection{Coherence analysis details.}

In the main paper, we perform two analyses: one to calculate the deviation from the expected number of holes and the other to calculate the border F1. We use these metrics to qualitatively evaluate coherence based on the fact that coherent prediction should: (i) not contain sporadic holes in change masks (hole deviation metric) and (ii) be precise at borders.

To compute these metrics, we operate on the binary change mask produced by each method.

For the \textit{hole metric}, we apply the same procedure to the \emph{background}: we identify all background connected components and discard those that touch the image border, since such regions represent true background rather than holes. Among the remaining enclosed components, we keep only those whose area exceeds the minimum threshold of 10px, and their count forms the hole count for that prediction. The difference between this value and the ground-truth hole count (obtained as explained above, but with ground truth mask) yields the hole deviation.

For border F1 (BF1), we follow the standard F1 definition but only compute it for a 3-px-wide border around ground-truth binary masks.

Both metrics quantify structural coherence by penalising either unwanted perforation of change regions (holes) or imprecision at the border. Lower deviation indicates that a method produces more globally consistent change masks, while higher BF1 indicates better border precision.

\subsection{Related methods implementation details}
\label{asub:rel_impl}

For all models, we use the same data as in our case. We do adopt the normalisation and other model-specific settings for data processing.

We use the official code, hyperparameters, and weights provided by the authors for all evaluated remote sensing foundation models. The specific versions of the code used are as follows (repo + commit):

\begin{itemize}
    \item GFM~\cite{mendieta2023gfm}: \textcolor{weights}{GFM commit: 4dd248e8544b3b6a49f5173b0931d97a17a7f424}
    \item MTP~\cite{wang2024mtp}: \textcolor{weights}{MTP commit: 962f7fd8781c095eb26db65ead3016e666b6d417}
    \item SatDiFuser~\cite{jia2025satdifuser}: \textcolor{weights}{MTP commit: 962f7fd8781c095eb26db65ead3016e666b6d\-417}
\end{itemize}

Since foundation models lack a predefined, exact change-detection architecture, we adopt the authors' architecture code and load the weights as the \textit{encoder} into the BTC framework~\cite{rolih2025btc}. The configuration for MTP and GFM is the same as in~\cite{rolih2025btc}. For SatDiFuser, we use the default parameters and UPerNet decoder with simple feature difference, similar to BTC~\cite{rolih2025btc}.

We use official code, hyperparameters, and weights (where applicable) for all change detection methods. The following are repos and commits:
\begin{itemize}
    \item FCS-Diff~\cite{daudt2018fcn}:  \textcolor{weights}{fully\_convolutional\_change\_detection commit: 4dd83231f25\-319a7ebb16cbfa9912541ceabac9a}
    \item ChangeFormer~\cite{bandara2022changeFormer}:  \textcolor{weights}{ChangeFormer commit: afd1b7ed640aa265a2c730de9584\-16ae7356a2f9}
    \item SwinSUNet~\cite{zhang2022swinsunet}:  \textcolor{weights}{SwinSUNet commit: 721daf84238eda40fb49d626c21df4ed\-2246aa9e}
    \item GCD-DDPM~\cite{wen2024gcd-ddpm}: \textcolor{weights}{GCD commit: ecf2f25c55e849dc92d948e6ed0ed9ff05163b\-96}
    \item BiFA~\cite{zhang2024bifa}:  \textcolor{weights}{BiFA commit: 56cd0da461e5e4b0d6a9b4f3321f0a81a91d21b8}
    \item MaskCD~\cite{yu2024maskcd}: \textcolor{weights}{MaskCD commit: 31e3e15c50a81a369fc7fec2134b61fbedaa60\-05}
    \item ChangeMamba~\cite{chen2024changeMamba}: \textcolor{weights}{ChangeMamba commit: a91b82ee45059ce159f5f6f5d8e\-5818c33b84e68}
    \item HySCDG~\cite{benidir2025hyscdg}: \textcolor{weights}{HySCDG commit: 05db2154dc9f24ee650fb27285617abbf38d\-8a9e}, pretrained model weights from HF: \textcolor{weights}{Yanis236/FSC-Pretrained}
    \item DDPM-CD~\cite{nichol2021ddpm}: \textcolor{weights}{ddpm-cd commit: 4970792f65227958ffaa1de787649ce2c58\-39f12}
    \item BTC~\cite{rolih2025btc}: \textcolor{weights}{BTC-change-detection commit: db41090f2f26b84b2bd803b5177\-56a10f0805b2f}
    \item ChangeDino~\cite{cheng2025changedino}: \textcolor{weights}{ChangeDINO commit: 1870b2641b0eb83d367a484e53e023\-b578b26c1f}
\end{itemize}

We keep all hyperparameters the same as those set by the authors, except for the epoch count on SYSU and OSCD, where we perform some tuning to improve performance given the dataset size differences. 

Due to the unavailability of some code for SCD, we use results from the TaCo paper~\cite{guo2025taco}, the Change3D paper~\cite{zhu2025change3d}, and the UniChange paper~\cite{zhang2026unichange}.

\end{document}